\pdfoutput=1

\documentclass[11pt]{article}
\usepackage[dvipsnames]{xcolor}
\usepackage[final]{acl}

\usepackage{times}
\usepackage{latexsym}
\usepackage{enumitem}

\usepackage[T1]{fontenc}

\usepackage[utf8]{inputenc}

\usepackage{microtype}

\usepackage{inconsolata}

\usepackage{graphicx}
\usepackage{caption}
\usepackage{subcaption}
\usepackage{amsmath}

\usepackage{booktabs}
\usepackage{tabularx}
\usepackage{multirow}
\usepackage{placeins}
\usepackage{rotating}
\usepackage{makecell}

\usepackage{listings} 
\usepackage{tcolorbox}

\definecolor{pptyellow}{RGB}{215,215,2}
\definecolor{pptblue}{RGB}{0,255,255}
\definecolor{pptgreen}{RGB}{0,176,80}

\usepackage{authblk}
\usepackage{footmisc}
\title{Enhancing Hallucination Detection via Future Context}

\author{
  \textbf{Joosung Lee}$^1$ \quad
  \textbf{Cheonbok Park}$^{1,3}$ \quad
  \textbf{Hwiyeol Jo}$^1$ \quad
  \textbf{Jeonghoon Kim}$^{1,3}$\\
  \textbf{Joonsuk Park}$^{1,2,4}$\textsuperscript{$\dagger$} \quad
  \textbf{Kang Min Yoo}$^1$\textsuperscript{$\dagger$}\\
  $^1$NAVER CLOUD\quad$^2$NAVER AI Lab\quad$^3$KAIST\quad$^4$University of Richmond\\
  \texttt{\{rung.joo,kangmin.yoo\}@navercorp.com, park@joonsuk.org}
}

\begin{document}
\maketitle
\begin{abstract}
Large Language Models (LLMs) are widely used to generate plausible text on online platforms, without revealing the generation process.
As users increasingly encounter such black-box outputs, detecting hallucinations has become a critical challenge.
To address this challenge, we focus on developing a hallucination detection framework for black-box generators.
Motivated by the observation that hallucinations, once introduced, tend to persist, we sample future contexts.
The sampled future contexts provide valuable clues for hallucination detection and can be effectively integrated with various sampling-based methods.
We extensively demonstrate performance improvements across multiple methods using our proposed sampling approach.
\end{abstract}

\renewcommand{\thefootnote}{\fnsymbol{footnote}}
\footnotetext[2]{Corresponding authors.}

\section{Introduction}
Large Language Models (LLMs) are widely used for various text generation tasks due to their fluency and plausibility.
However, these models frequently exhibit overconfidence or generate incorrect information—a phenomenon known as hallucination—which remains a critical challenge in leveraging LLMs effectively across applications~\cite{Ji_2023, sahoo-etal-2024-comprehensive}.

\begin{figure}[t]
    \centering
    \includegraphics[width=0.7\linewidth]{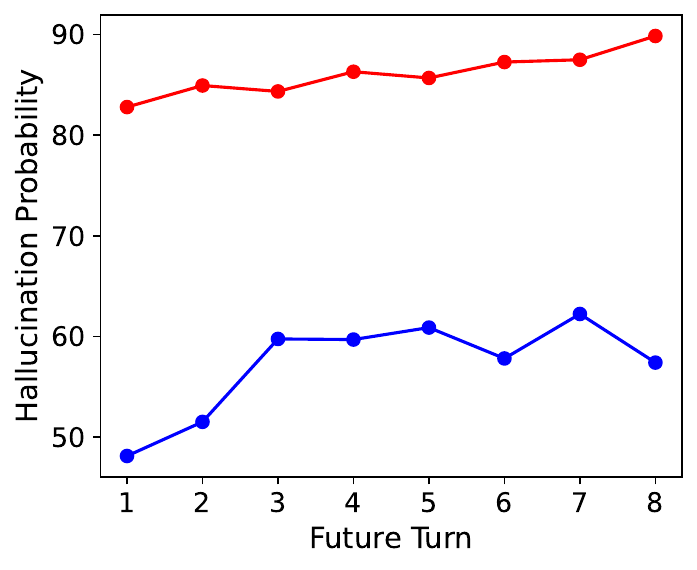}
    \caption{Temporal influence of hallucinated sentences in SelfCheckGPT dataset (based on gold labels). 
    The plot shows the probability that the $i$-th future sentence is hallucinated, given that the current sentence is non-hallucinated (\textcolor{blue}{blue}) or hallucinated (\textcolor{red}{red}).}
    \label{fig:hall_stats}
\end{figure}

LLM-based hallucination detection methods largely fall into two main categories: those based on uncertainty~\cite{stitch, enhancing-uncertainty, ccp} and those based on sampling strategies~\cite{selfcheckgpt, sac, selfcont, interrogatellm}.
However, existing approaches suffer from practical issues such as the unavailability of the generator model’s logits in many real-world scenarios and increased sampling costs. 
For example, in online platforms such as blog posts, it is often difficult to determine how the content was generated.
Moreover, API-based generators may become inaccessible over time due to updates or deprecation. 
Consequently, uncertainty-based methods that rely on generator-specific token-level logits are impractical. 

Retrieval-based approaches are effective for verifying general factual information; however, they face significant limitations when internal documents or proprietary knowledge bases are required, as access to these resources is frequently restricted or unavailable. Moreover, retrieval-based methods often struggle to detect logical hallucinations and internal inconsistencies, limiting their effectiveness. 
Indeed, \citet{selfcont} report that 35.2\% of self-contradictory hallucinations generated by ChatGPT remain undetectable through retrieval-based verification, suggesting that retrieval-based approaches are not always effective.
Consequently, we focus on sampling-based hallucination detection methods that operate in a fully black-box setting without retrieval.
In contrast to approaches that rely on internal model signals such as attention (\citet{chuang-etal-2024-lookback}),
our method leverages future context, making it applicable to diverse and realistic scenarios.

\begin{figure*}[t]
    \centering
    \includegraphics[width=1.0\linewidth]{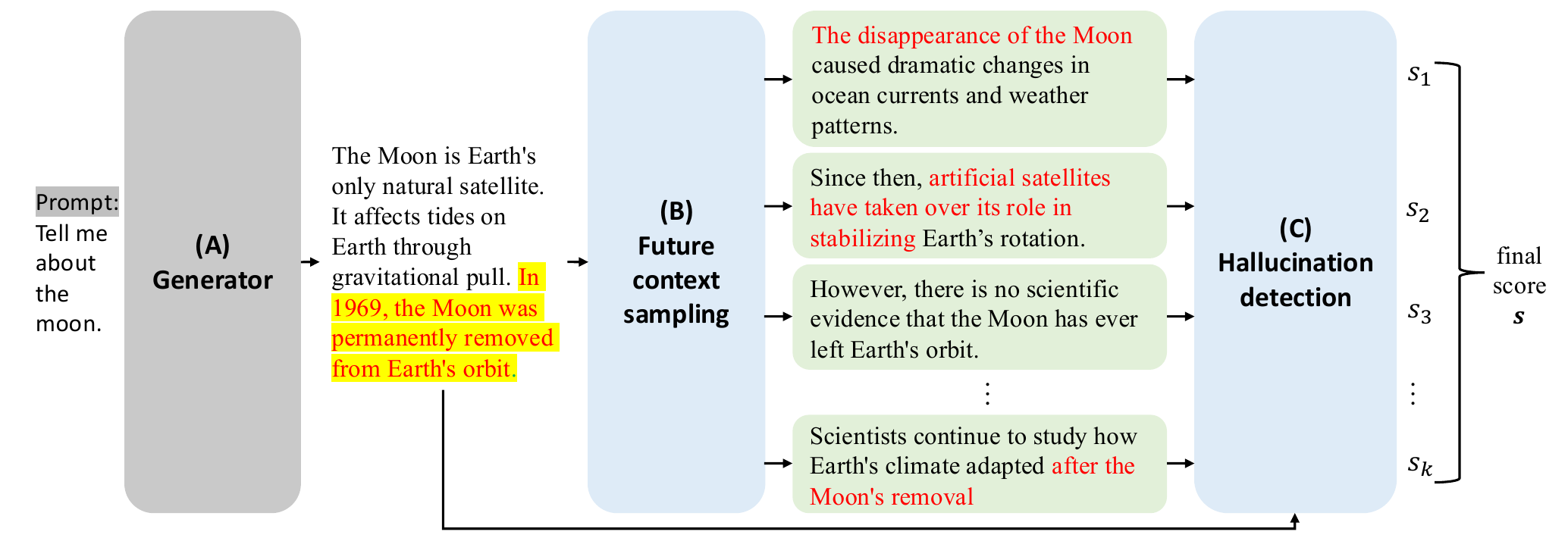}
    \caption{Our proposed hallucination detection pipeline. (A) A black-box generator produces context–response pairs without exposing its internal generation process. (B) Future context sampling: an instruction-tuned LLM generates possible next sentences. (C) Hallucination detection: each sentence is scored using the context, response, and the sampled future context.
    When the target sentence (\textcolor{pptyellow}{yellow highlight}) is hallucinated (\textcolor{red}{red text}), sampled future contexts are more likely to contain hallucinated information. We propose to leverage this to detect hallucination in the target sentence more reliably.
    }
    \label{fig:overview}
\end{figure*}

To cope with these issues, we propose to generate future context and leverage it for hallucination detection in the target sentence.
Our method is motivated by the observed snowball effect~\cite{snowball}, where hallucinated sentences are often correlated with hallucinations in future context.
For instance, in answering yes/no questions, if the first token of a response is \textit{yes}, subsequent sentences tend to support this claim, regardless of correctness.
Figure~\ref{fig:hall_stats}, using the SelfCheckGPT dataset, shows that a hallucinated sentence increases the likelihood of hallucinations in subsequent sentences compared to a factual sentence.
Furthermore, hallucination errors increasingly propagate to later sentences.
Motivated by these observations, we leverage sampled future contexts as clues to detect hallucinations in the target sentence.
Section~\ref{sec:how_future} confirms that hallucinated current sentences increase the likelihood of hallucinations in future sentences, making the future context a useful signal for detecting hallucinations in the current sentence.

We combine future context with two previous sampling-based methods, \textsc{SelfCheckGPT} and \textsc{SC}~\cite{selfcont}, as well as our proposed baseline, \textsc{Direct}.
\textsc{Direct} is a simple baseline method that explicitly determines sentence hallucination by directly leveraging the detector LLM's internal knowledge and reasoning abilities.
Through extensive experiments using three detector models (LLaMA 3.1, Gemma 3, Qwen 2.5), we demonstrate that incorporating sampled future context consistently improves hallucination detection performance across various configurations, including the number of sampled future sentences and lookahead turns.
Additionally, we demonstrate that combining future context with the existing sampling clues used in \textsc{SelfCheckGPT} effectively reduces sampling costs.

Our main contributions are as follows:
\begin{itemize}[nosep, leftmargin=1em]
    \item We propose leveraging future context to enhance sampling-based hallucination detection, achieving state-of-the-art performance when combined with existing methods.
    
    \item Our method is generator-agnostic and can be easily combined with existing frameworks, providing high extensibility across various environments.

    \item Through extensive analyses, we demonstrate that enriching future context by increasing samples substantially improves hallucination detection performance.
\end{itemize}

\section{Related Work}
\label{sec:related_hallucination_detection}
One line of research detects hallucinations by leveraging sampling-based strategies.
\textsc{\textbf{SelfCheckGPT}}~\cite{selfcheckgpt} detects hallucinations by generating multiple context-response pairs and evaluating whether they support the original output. If the sampled pairs provide consistent evidence for the original response, it is deemed factual; otherwise, it is flagged as a hallucination.
\textsc{\textbf{SAC$^{3}$}}~\cite{sac} enhances sampling-based detection by paraphrasing the input and leveraging multiple models to produce diverse outputs. This multi-perspective generation improves the ability to uncover hallucinations that may escape traditional consistency checks.
\textsc{\textbf{SC}} (Self-Contradiction Detection)~\cite{selfcont} addresses hallucinations by identifying internal contradictions within generated responses. It regenerates each sentence based on its extracted context and compares it with the original to detect inconsistencies. 

Another line of research focuses on token-level uncertainty to assess factuality. \textsc{\textbf{Stitch}}~\cite{stitch} identifies hallucinations by analyzing low-confidence tokens. It shows that the minimum token probability provides an effective signal for detecting factual errors.
\textsc{\textbf{CCP}} (Claim Conditioned Probability)~\cite{ccp} measures uncertainty at both token and claim levels using model-generated alternative tokens. These are assessed via natural language inference to estimate the overall factuality of the claim.

We propose a novel sampling-based approach that leverages future context to enhance hallucination detection. 
By incorporating future context as a clue in existing methods, our approach improves performance in black-box generator settings. 

\section{Approach}
Figure~\ref{fig:overview} is an overview of our method.  
Since we assume no access to the generator, the detector LLM need not match the generator.

\subsection{Future Context Sampling}
We use an instruction-tuned LLM as a future context sampler, prompting the model to generate the next sentence.
When generating future context that extends beyond the immediate next sentence, we found it more effective to instruct the model to generate multiple sentences simultaneously, rather than generating them sequentially.
A future context is defined as the set of sentences generated through a single sampling path.

\subsection{Incorporating Future Context into Hallucination Detection}
We explore the integration of future context into various sampling-based methods for hallucination detection. 
While the effective strategies for utilizing future context may differ across methods, we adopt a simple and unified approach: appending future context directly to the prompt.

\paragraph{\textsc{SelfCheckGPT} with Future Context.}
First, we incorporate future context into \textsc{SelfCheckGPT}, a method that detects hallucinations by evaluating the consistency between the original response and multiple sampled alternative context-response pairs.
The core idea of \textsc{SelfCheckGPT} is that factual information tends to be consistently present across different sampled responses, whereas hallucinated information tends to vary or contradict.
By appending future context to the alternative context-response pairs, we expand the clues available to the LLM for detecting hallucinations.

\paragraph{\textsc{SC} with Future Context.}
Second, we incorporate future context into \textsc{SC}, which samples alternative current-turn responses to identify inconsistencies.
We omit the description field in the original \textsc{SC} prompt for consistency across datasets lacking this field.
In settings where future context is used, we instead substitute the description with a future context.

\paragraph{\textsc{Direct} with Future Context.}
Finally, we apply future context to our proposed baseline method \textsc{Direct}. 
Previous works~\cite{LLmKnow1, LLmKnow2} demonstrate that LLMs exhibit a form of self-awareness, suggesting they can leverage their internal knowledge for hallucination detection. 
Motivated by this, we introduce \textsc{Direct}, explicitly prompting the detector to determine hallucinations.
\textsc{Direct} determines hallucination without relying on complex probability estimations or sampling techniques.
This approach relies solely on the internal knowledge and reasoning capabilities of the detector LLM, serving as a straightforward and effective starting point for leveraging the model's contextual understanding and factual grounding.
This approach allows for precise control over the key elements involved in hallucination detection, making it well-suited for comprehensively analyzing the effects of future context.

\textsc{Direct} poses a straightforward binary (\textit{yes}/\textit{no}) question to the model, asking whether a given sentence is accurate. 
The prompt consolidates information, aiding the model's internal knowledge-based hallucination detection.
In this sense, this approach serves as a strong baseline and a practical starting point for hallucination detection using LLMs.
All prompt templates are provided in Appendix~\ref{app:template_prompts}.

\subsection{Hallucination Score}
In the sampling-based methodology, following \textsc{SelfCheckGPT}, we pair the target sentence with multiple sampled clues.
In \textsc{Direct}, each clue corresponds to a sampled future context. 
For \textsc{SelfCheckGPT} and \textsc{SC}, each clue is formed by combining their originally sampled outputs with the sampled future context.
Each sentence-clue pair is individually evaluated for hallucination: if the response is judged to be a hallucination given the clue, it is assigned a score of 0; if not, it receives a score of 1; and if the judgment is indeterminate (i.e., N/A), a score of 0.5 is assigned.
Finally, the scores across all pairs are averaged to obtain the hallucination score for the sentence.

\begin{table*}[t]
    \centering
    \resizebox{0.9\textwidth}{!}{
    \begin{tabular}{ll|cccccc|c}
    \toprule
    \multirow{2}{*}{Detector} & \multirow{2}{*}{Method} & \multicolumn{6}{c|}{Dataset} & \multirow{2}{*}{Average} \\
    & & SelfCheckGPT & SC-ChatGPT & SC-GPT4 & SC-LLaMA & SC-Vicuna & True-False & \\    
    \midrule
    \multirow{6}{*}{LLaMA 3.1}
    & \textsc{Direct}         & 62.1 & 66.4 & 63.8 & 65.9 & 68.0 & 87.2 & 68.9 \\
    & \textsc{Direct}+f       & \textbf{65.3} & \textbf{69.0} & \textbf{68.0} & \textbf{67.1} & \textbf{68.4} & \textbf{88.5} & \textbf{71.1}  \\
    \cmidrule{2-9}
    & \textsc{SelfCheckGPT}      & 67.0 & 71.2 & 76.0 & 68.1 & 69.6 & 89.2 & 73.5 \\
    & \textsc{SelfCheckGPT}+f    & \textbf{67.3} & \textbf{74.9} & \textbf{76.4} & \textbf{69.8} & \textbf{70.2} & \textbf{90.3} & \textbf{74.8}  \\
    \cmidrule{2-9}
    & \textsc{SC}        & 53.2 & 62.5 & 65.8 & 67.1 & 67.3 & 78.1 & 65.7 \\
    & \textsc{SC}+f      & \textbf{57.7} & \textbf{69.2} & \textbf{73.3} & \textbf{71.2} & \textbf{71.4} & \textbf{81.8} & \textbf{70.8}  \\
    \midrule
    \multirow{6}{*}{Gemma 3}
    & \textsc{Direct}        & 53.7 & 52.5 & 53.4 & 55.6 & 56.6& 87.9 & 59.9   \\
    & \textsc{Direct}+f     & \textbf{56.6} & \textbf{58.0} & \textbf{58.0} & \textbf{58.1} & \textbf{57.5} & \textbf{90.8} & \textbf{63.2}   \\
    \cmidrule{2-9}
    & \textsc{SelfCheckGPT}    & 49.3 & 68.8 & 71.9 &68.7 &67.7 &89.7  &69.4    \\
    & \textsc{SelfCheckGPT}+f  & \textbf{58.3} & \textbf{71.0} & \textbf{72.5} & \textbf{71.2} & \textbf{70.1} & \textbf{91.3} & \textbf{72.4}   \\
    \cmidrule{2-9}
    & \textsc{SC}      & 52.9 & 61.1 & 62.3 & 64.2& 65.7& 86.1 & 65.4   \\
    & \textsc{SC}+f    & \textbf{55.2} & \textbf{65.6} & \textbf{65.3} & \textbf{66.2} & \textbf{67.7} & \textbf{86.1} & \textbf{67.7}   \\
    \midrule
    \multirow{6}{*}{Qwen 2.5}
    & \textsc{Direct}         & 60.6 & 63.0 & 67.9 & 64.8 & 65.5 & 82.6 & 67.4 \\
    & \textsc{Direct}+f       & \textbf{62.9} & \textbf{65.8} & \textbf{69.1} & \textbf{67.0} & \textbf{65.6} & \textbf{85.7} & \textbf{69.4}  \\
    \cmidrule{2-9}
    & \textsc{SelfCheckGPT}      & 51.7 & 64.4 & 67.8 & 63.5 & 65.7 & 87.4 & 66.7 \\
    & \textsc{SelfCheckGPT}+f    & \textbf{51.7} & \textbf{65.5} & \textbf{70.3} & \textbf{65.4} & 65.1 & 85.9 & \textbf{67.3}  \\
    \cmidrule{2-9}
    & \textsc{SC}        & 50.5 & 50.7 & 50.6 & 52.3 & 56.0 & 69.3 & 54.9 \\
    & \textsc{SC}+f      & \textbf{54.2} & \textbf{59.5} & \textbf{60.5} & \textbf{60.0} & \textbf{60.2} & \textbf{70.2} & \textbf{60.8} \\
    \bottomrule
    \end{tabular}
    }
    \caption{Hallucination detection performance (AUROC) of \textsc{Direct}, \textsc{SelfCheckGPT}, and \textsc{SC} with and without future context (+f) across different LLM detectors.
    Bold numbers indicate performance improvements when future context (+f) are used.}
    \label{tab:main_table}
\end{table*}

\section{Datasets}
\label{sec:dataset}
Previous studies typically evaluate hallucination detection methods using data labeled by their respective generators, making it difficult to compare performance directly due to differences in generators, detectors, and dataset formats.
Furthermore, existing datasets differ in the types of hallucinations they address, with SelfCheckGPT and SC datasets targeting logical hallucinations and True-False focusing on factual hallucinations. 
To address this limitation, we standardize the dataset format and conduct unified evaluations across various methods. 
As our approach assumes a black-box generator setting, we do not require detectors tailored to specific generators, enabling diverse LLM-based detectors.
Examples are provided in Table~\ref{tab:example}.

SelfCheckGPT dataset~\cite{selfcheckgpt} is constructed using GPT-3 (text-davinci-003)~\cite{GPT3}, which is used to sample responses to given questions. 
Two annotators label each response as either accurate, minor inaccurate, or major inaccurate. If annotators disagree, the worse label is chosen.
We treat responses labeled as minor or major inaccurate as hallucinations.
The context used for response generation consists of model-generated sentences.

SC dataset~\cite{selfcont} is constructed using various LLMs, such as ChatGPT (gpt-3.5-turbo-0301)~\cite{openai2023chatgpt}, GPT4 (gpt-4-0314)~\cite{openai2024gpt4technicalreport}, LLaMA2-70B-Chat~\cite{touvron2023llama2openfoundation}, and Vicuna-13B~\cite{vicuna2023}, which generate responses based on the given question and context. An initial response is first generated, followed by an additional sampled response. Three annotators then assess whether a self-contradiction exists between the two responses. If a self-contradiction is identified, at least one of the responses must contain hallucinated content~\cite{dowden2011logical, selfcont}. 
This does not necessarily imply that the initial response is incorrect. 
Nevertheless, \citet{selfcont} has shown that removing all conflicting information from both sentences can improve overall response quality.
Therefore, despite certain limitations, we consider the initial response to be hallucinated when a self-contradiction is detected, in order to enable broader benchmarking.
We define a concept based on the string following the term ‘about’ in a given question.
Unlike SelfCheckGPT, where context is constructed from model-generated text, the SC dataset uses gold responses as context.
For convenience, we refer to this version of the SC dataset as ‘SC-\{LLM\}’.

True-False dataset~\cite{true_false} comprises individual sentences labeled as either true or false. Sentences labeled as false are considered hallucinations.
As this dataset does not provide prompts, we constructed prompts following the same format used in SC dataset.
The concept is extracted from each sentence using an LLM, and only sentences that contain the exact concept string are used for evaluation.
Moreover, to ensure fairness, the LLM used for concept extraction is different from the one used as the hallucination detector.
Table~\ref{tab:dataset_stats} presents dataset statistics\footnote{The reconstructed datasets will be released upon acceptance.}.

\section{Experiments}
\label{sec:experiments}
We employ three instruction-tuned LLM detectors: LLaMA 3.1-8B~\cite{llama}, Gemma 3-12B~\cite{gemma3}, and Qwen 2.5-7B~\cite{qwen}. 
By sampling future context from these detectors and integrating it with various methods, we demonstrate improved hallucination detection performance. 
We evaluate using AUROC as our primary metric because our dataset is not significantly imbalanced and AUROC is commonly adopted in related research. For completeness, we also report AUCPR results in Table~\ref{tab:aucpr}.

We use top-k sampling ($k=30$) to generate future and current sentences, and follow the original generation configurations for alternative context–response pairs in \textsc{SelfCheckGPT}.
Generation terminates when an end-of-sequence token is produced, and the resulting text is segmented into sentences using SpaCy~\footnote{https://spacy.io}.
Finally, we use greedy decoding when the detector generates responses to prompts for hallucination detection.

\begin{figure*}[!t]
    \centering
    \renewcommand{\arraystretch}{1.2}
    \setlength{\extrarowheight}{2pt}
    \resizebox{1.0\textwidth}{!}{
        \begin{tabular}{c ccc}
            \raisebox{3.5em}{\rotatebox{90}{\textsc{SelfCheckGPT}}} &
            \begin{subfigure}{0.32\textwidth}
                \includegraphics[width=\textwidth]{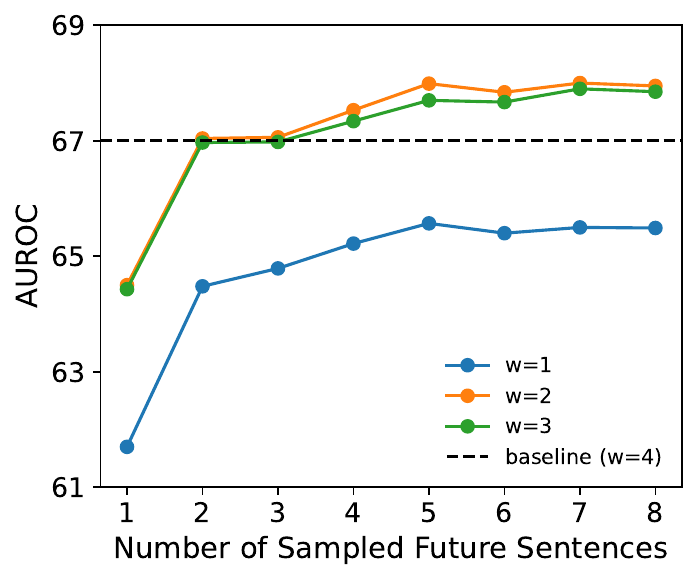}
                \caption{Dataset: SelfCheckGPT}
                \label{fig:selfcheckgpt_selfcheckgpt}
            \end{subfigure} &
            \begin{subfigure}{0.32\textwidth}
                \includegraphics[width=\textwidth]{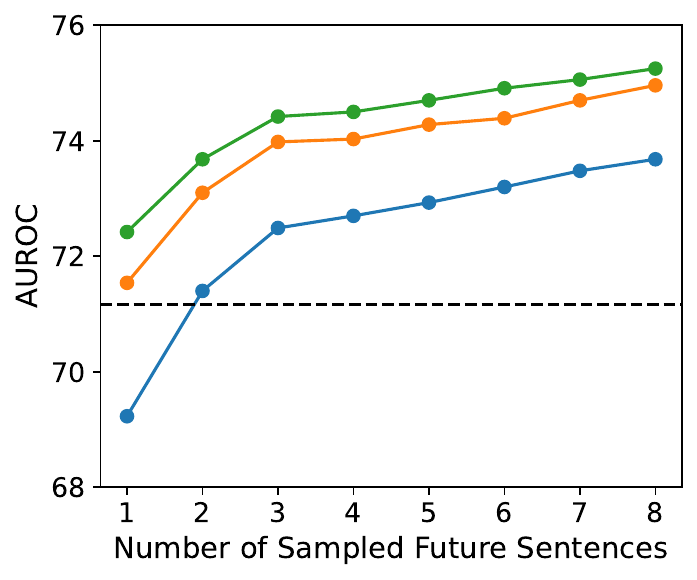}
                \caption{Dataset: SC-ChatGPT}
                \label{fig:selfcheckgpt_chatgpt}
            \end{subfigure} &
            \begin{subfigure}{0.32\textwidth}
                \includegraphics[width=\textwidth]{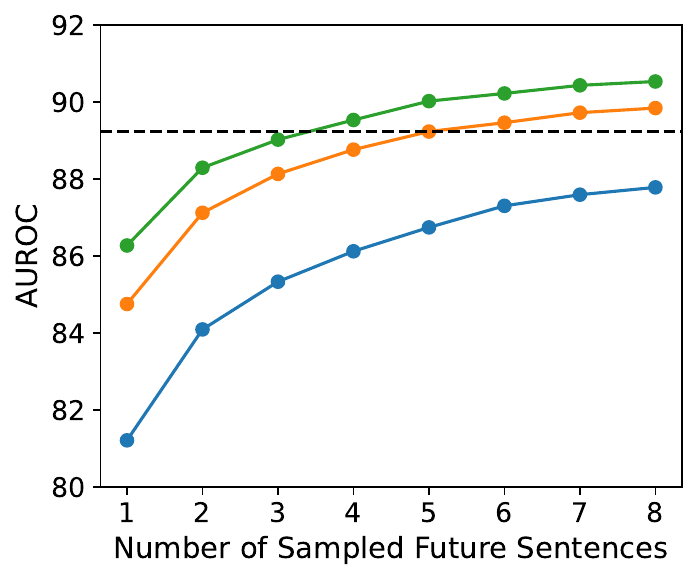}
                \caption{Dataset: True-False}
                \label{fig:selfcheckgpt_truefalse}
            \end{subfigure}
            \\
            \raisebox{7em}{\rotatebox{90}{\textsc{SC}}} &
            \begin{subfigure}{0.32\textwidth}
                \includegraphics[width=\textwidth]{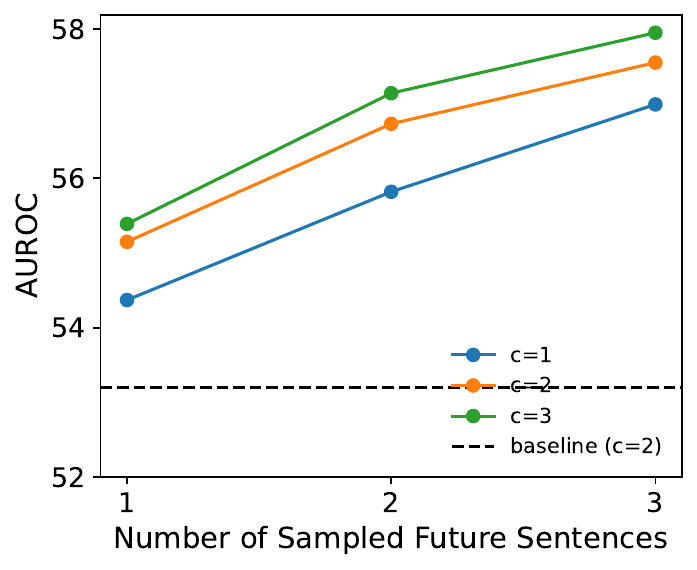}
                \caption{Dataset: SelfCheckGPT}
                \label{fig:sc_selfcheckgpt}
            \end{subfigure} &
            \begin{subfigure}{0.32\textwidth}
                \includegraphics[width=\textwidth]{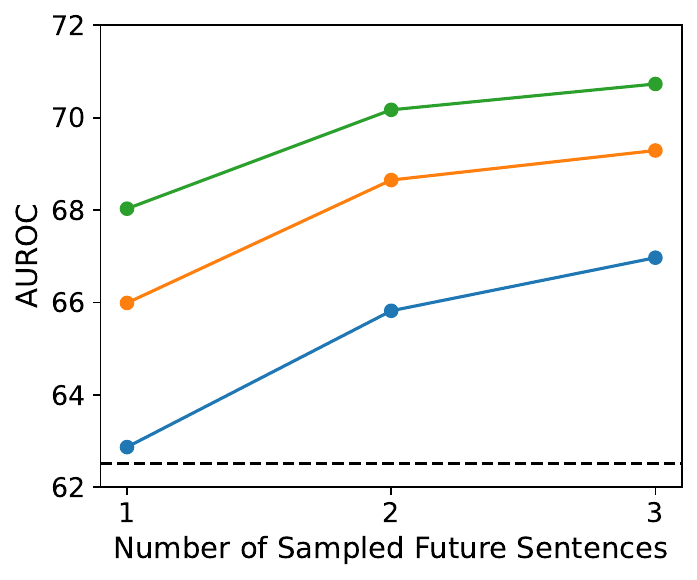}
                \caption{Dataset: SC-ChatGPT}
                \label{fig:sc_chatgpt}
            \end{subfigure} &
            \begin{subfigure}{0.32\textwidth}
                \includegraphics[width=\textwidth]{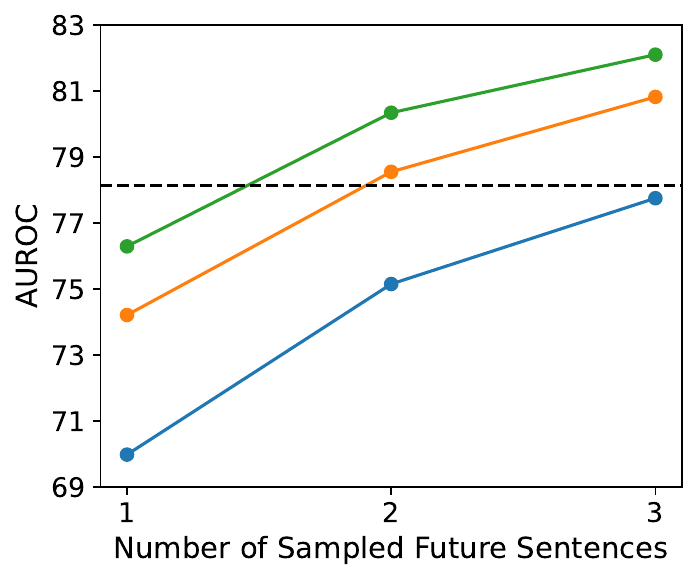}
                \caption{Dataset: True-False}
                \label{fig:sc_truefalse}
            \end{subfigure}
        \end{tabular}
    }
    \caption{AUROC of \textsc{SelfCheckGPT} and \textsc{SC} with future contexts (detector: LLaMA 3.1).
    The first and second rows show performance improvements for \textsc{SelfCheckGPT} and \textsc{SC}, respectively, when incorporating future contexts.
    In both cases, performance further increases as the number of sampled future contexts grows.}
    \label{fig:selfcheck_w_sc_c}
    \vspace{-3mm}
\end{figure*}

\begin{figure}[t]
    \centering    \includegraphics[width=1.0\linewidth]{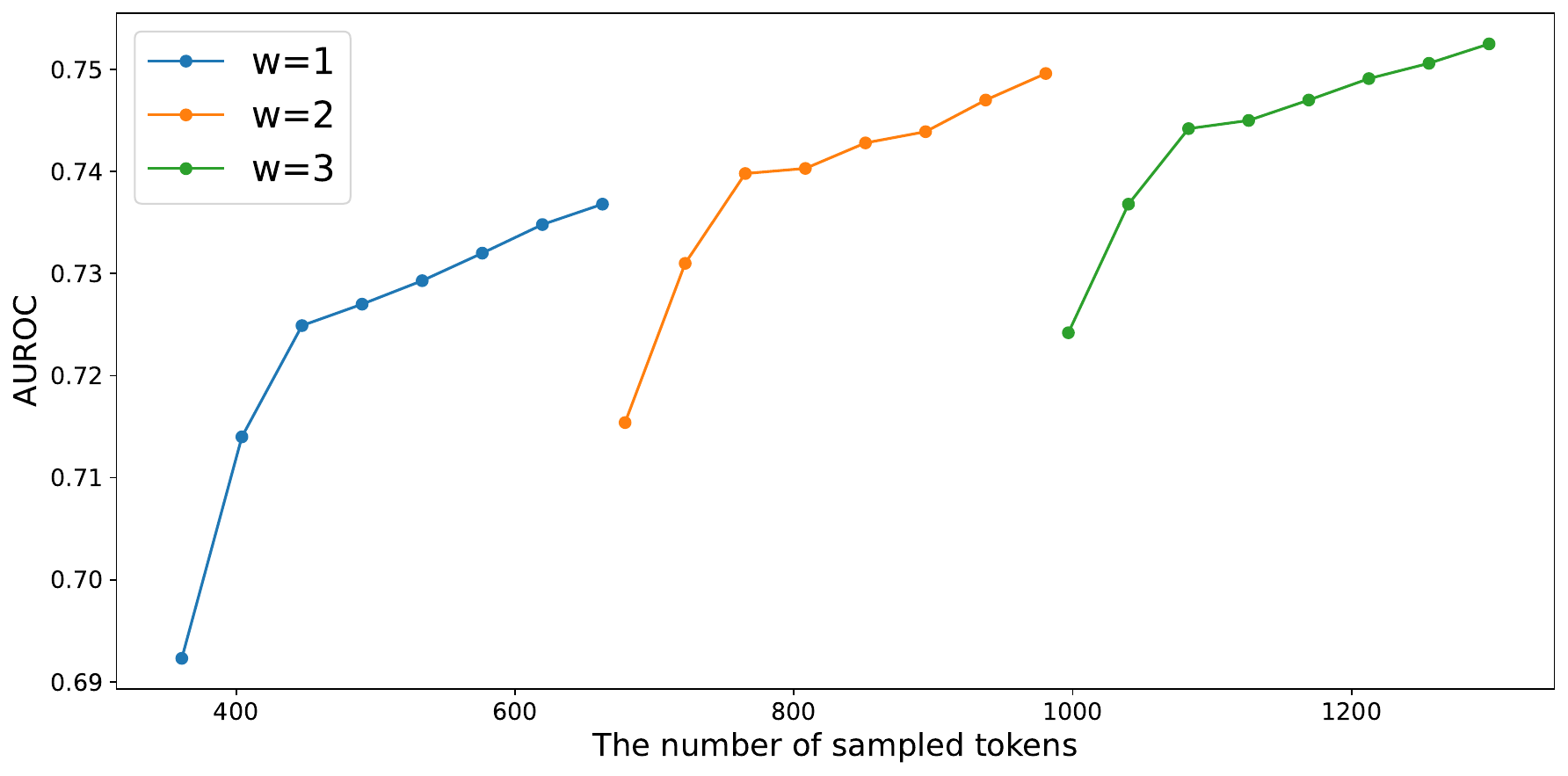}
    \caption{AUROC vs. sampled token consumption using the \textsc{SelfCheckGPT}+f on the SC-ChatGPT (Detector: LLaMA 3.1). 
    }
    \label{fig:token_performance}
\end{figure}

\subsection{Future Context Improves Performance}
We investigate performance improvements by incorporating future context into sampling-based methods.  
Table~\ref{tab:main_table} shows the performance of \textsc{Direct}, \textsc{SelfCheckGPT}, and \textsc{SC}—each evaluated with and without future context—across different LLM detectors.  
In \textsc{SelfCheckGPT}, we use $w = 4$ alternative context-response pairs, where $w$ denotes the number of context variations paired with the same response.
In \textsc{SC}, we use $c = 2$, representing the number of sampled responses for the current sentence.
In each of the future-augmented variants—\textsc{Direct}+f, \textsc{SelfCheckGPT}+f, and \textsc{SC}+f—we sample four one-turn-ahead future sentences and combine them with the original prompts.
Incorporating future context improves performance in most cases.

For LLaMA 3.1 and Gemma 3, \textsc{SelfCheckGPT} achieves the best performance, whereas for Qwen 2.5, \textsc{Direct}+f performs best.  
Sampling-based methods generally benefit from increased samples, leading to improved performance~\cite{selfcheckgpt}.
Table~\ref{tab:sampled_tokens} presents the average token counts of sampled outputs for each LLM.  
Since \textsc{SelfCheckGPT} samples entire alternative context-response pairs, it requires more tokens compared to methods sampling only future or current sentences.  
Thus, even with equal samples, \textsc{SelfCheckGPT} incurs higher computational costs.  
For comprehensive analysis, we extend sampling targets ($w, c$) with future sentences across various parameter combinations.

\begin{figure}[t]
    \centering    
    \includegraphics[width=1.0\linewidth]{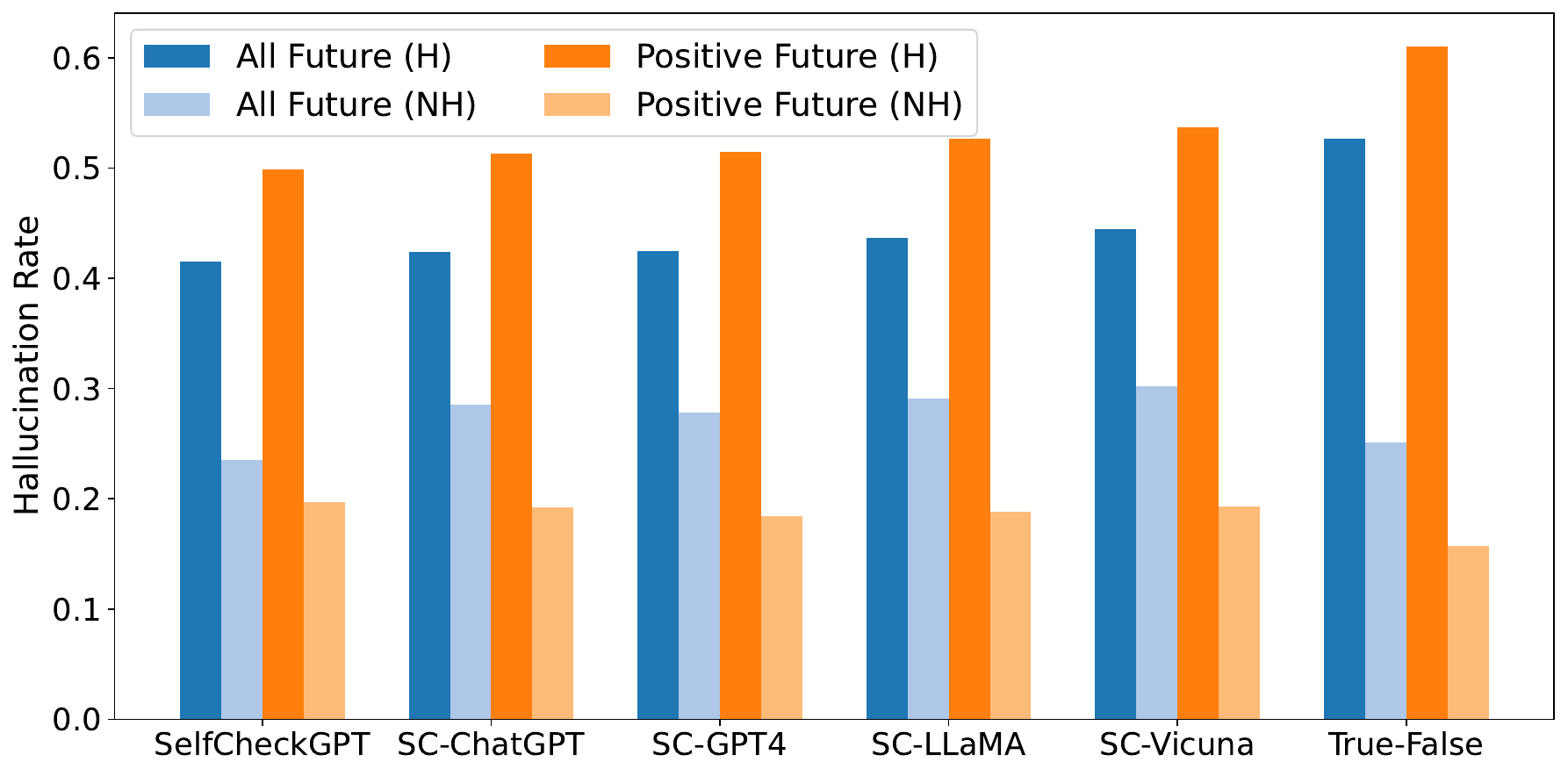}
    \caption{Hallucination rates of future sentences (as shown in the top row of Figure~\ref{fig:direct_s_t_ind}, LLaMA 3.1), conditioned on the hallucination label of the current sentence.
    \textbf{All Future (H)}: The proportion of hallucinated future sentences when the current sentence is hallucinated.
    \textbf{All Future (NH)}: The proportion of hallucinated future sentences when the current sentence is non-hallucinated.
    \textbf{Positive Future (H)}: Proportion of hallucinated future sentences aiding correct classification of hallucinated current sentences.
    \textbf{Positive Future (NH)}: Proportion of hallucinated future sentences aiding correct classification of non-hallucinated current sentences.
    }
    \label{fig:analysis_future_h}
    \vspace{-4mm}
\end{figure}

\begin{figure}[t]
    \centering
    \includegraphics[width=1.0\linewidth]{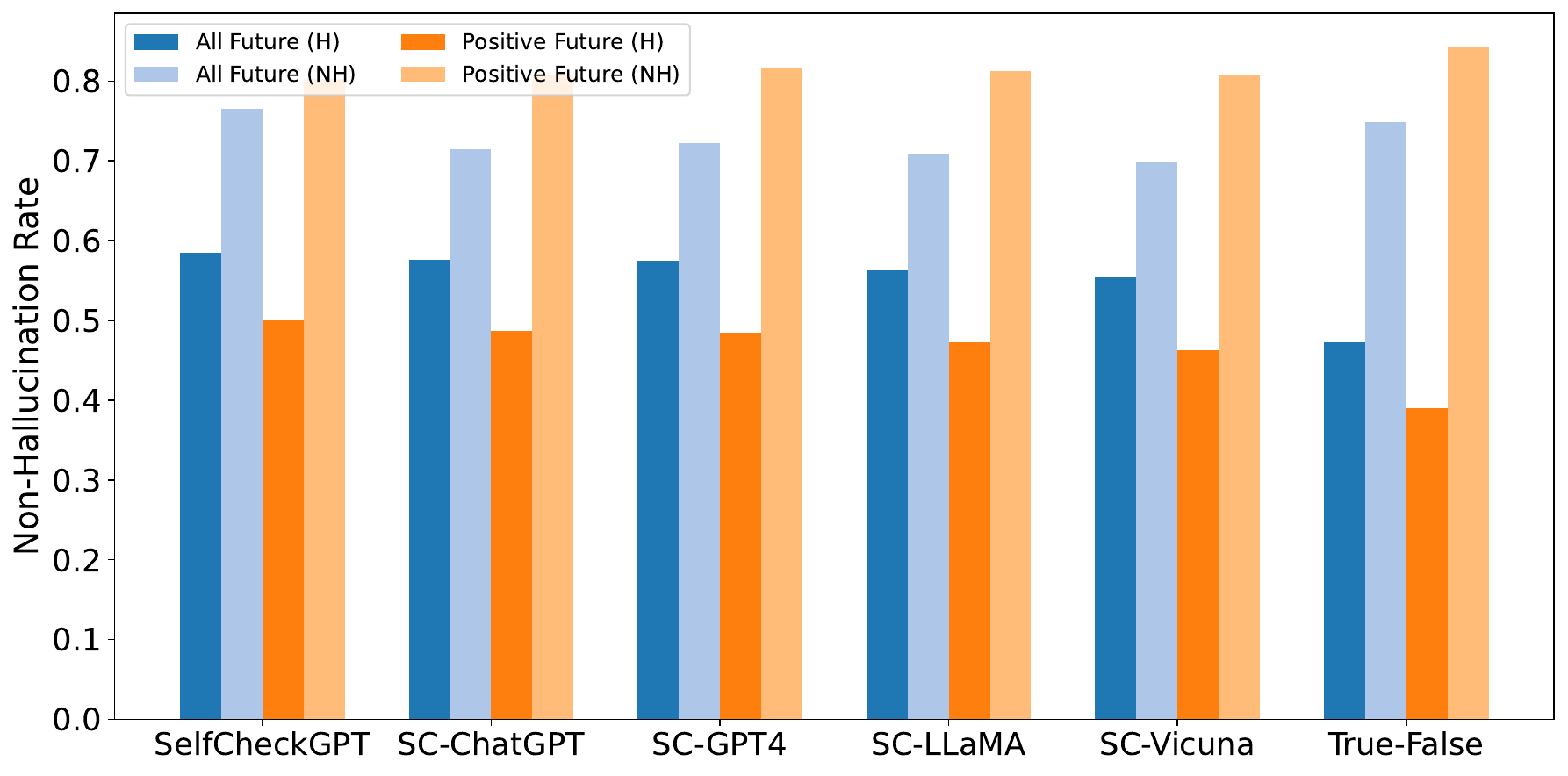}
    \caption{Non-Hallucination rates of future sentences (as shown in the top row of Figure~\ref{fig:direct_s_t_ind}, LLaMA 3.1) conditioned on the hallucination label of the current sentence.
    \textbf{All Future (H)}: The proportion of non-hallucinated future sentences when the current sentence is hallucinated.
    \textbf{All Future (NH)}: The proportion of non-hallucinated future sentences when the current sentence is non-hallucinated.
    \textbf{Positive Future (H)}: Proportion of non-hallucinated future sentences aiding correct classification of hallucinated current sentences.
    \textbf{Positive Future (NH)}: Proportion of non-hallucinated future sentences aiding correct classification of non-hallucinated current sentences.    
    }
    \label{fig:analysis_future_nh}
    \vspace{-4mm}
\end{figure}

Figure~\ref{fig:selfcheck_w_sc_c} (top row) shows the performance of \textsc{SelfCheckGPT} as the number of sampled \( w \) and \( s \) varies, with \( t  = 1 \), where $s$ is the number of future sentence samples per turn and $t$ is the number of future turns. 
Figure~\ref{fig:token_performance} further analyzes the relationship between AUROC and the number of sampled tokens across different $(w, s)$. The results indicate that reducing $w$ while increasing $s$ can be more token-efficient than solely increasing $w$, demonstrating cases where comparable or superior performance is achieved with fewer tokens. 
Since the number of sampled tokens significantly impacts inference speed during both response generation and hallucination detection, these findings support the cost-effectiveness of leveraging future contexts.
Additional analyses regarding the number of inference steps are provided in Appendices~\ref{app:agg_future} and~\ref{app:self_consistenccy}.

Figure~\ref{fig:selfcheck_w_sc_c} (bottom row) illustrates the performance of \textsc{SC} under various combinations of sampled \( c \) and \( s \).  
As with \textsc{SelfCheckGPT}, increasing $s$ alongside $c$ generally improves performance.
While incorporating $s$ does not reduce token costs (as $c$ and $s$ are similar in token length), combining $s$ with a fixed total sample count often improves performance.
This observation suggests the potential benefit of leveraging future context through the inclusion of \( s \).
Additional experimental results are provided in Figures~\ref{fig:selfcheck_w_sc_c2}.

Additionally, Appendix~\ref{app:retreival_methods} explores the potential effectiveness of incorporating future context into retrieval-based methods. 
Appendix~\ref{app:passage_level} further demonstrates its effectiveness in detecting passage-level hallucinations.

\begin{figure*}[t]
    \centering
    \renewcommand{\arraystretch}{1.2}
    \setlength{\extrarowheight}{2pt}
    \resizebox{1.0\textwidth}{!}{
        \begin{tabular}{>{\centering\arraybackslash}m{0.2cm} >{\centering\arraybackslash}m{0.34\textwidth} >{\centering\arraybackslash}m{0.34\textwidth} >{\centering\arraybackslash}m{0.34\textwidth}}
            \raisebox{5em}{\rotatebox[origin=c]{90}{Sampling Count}}
 &
            \begin{subfigure}[b]{\linewidth}
                \centering
                \includegraphics[width=\linewidth]{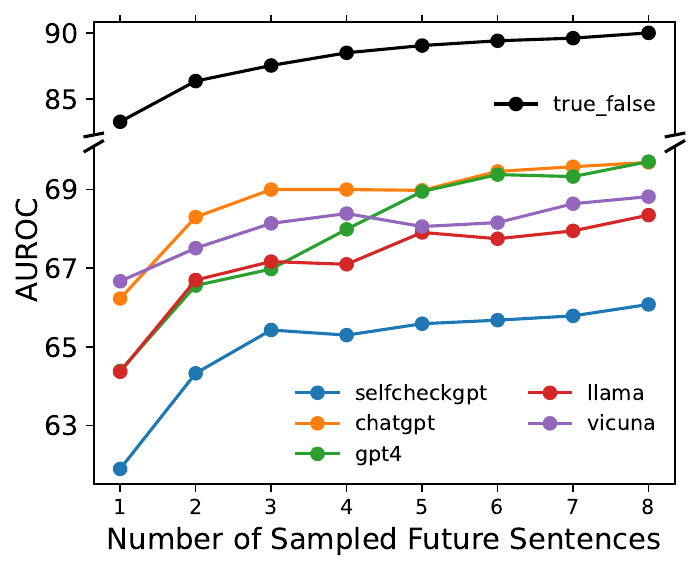}
                \caption{Detector: LLaMA 3.1}
                \label{fig:llama3.1_direct_s}
            \end{subfigure} &
            \begin{subfigure}[b]{\linewidth}
                \centering
                \includegraphics[width=\linewidth]{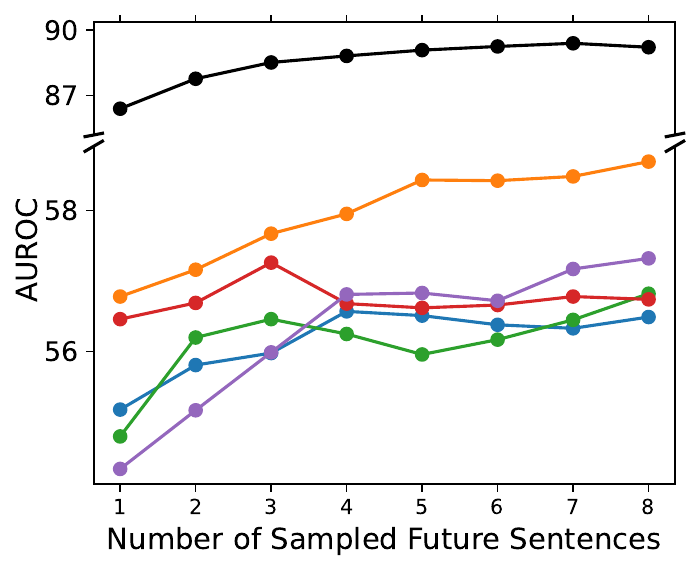}
                \caption{Detector: Gemma 3}
                \label{fig:gemma3_direct_s}
            \end{subfigure} &
            \begin{subfigure}[b]{\linewidth}
                \centering
                \includegraphics[width=\linewidth]{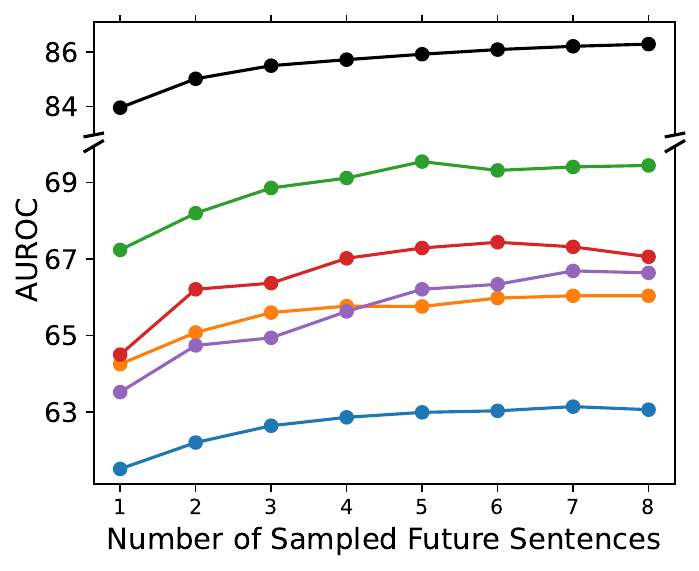}
                \caption{Detector: Qwen 2.5}
                \label{fig:qwen2.5_direct_s}
            \end{subfigure}
            \\[4ex] 
            \raisebox{5em}{\rotatebox[origin=c]{90}{Lookahead Turn}}&
            \begin{subfigure}[b]{\linewidth}
                \centering
                \includegraphics[width=\linewidth]{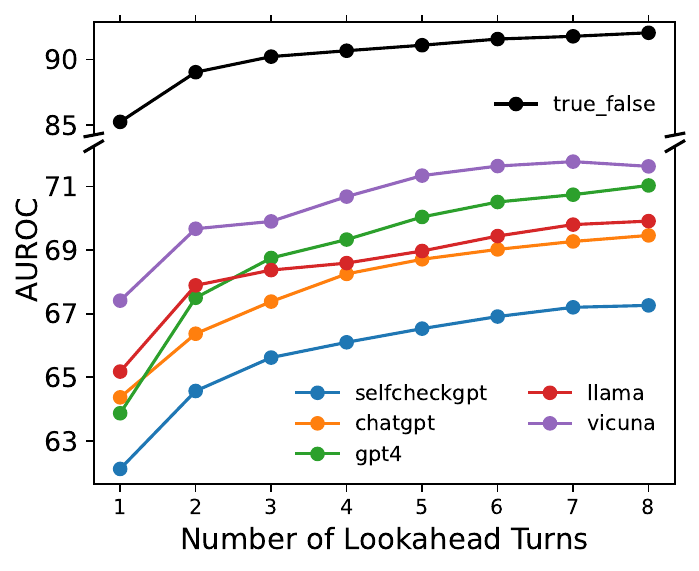}
                \caption{Detector: LLaMA 3.1}
                \label{fig:llama3.1_direct_t}
            \end{subfigure} &
            \begin{subfigure}[b]{\linewidth}
                \centering
                \includegraphics[width=\linewidth]{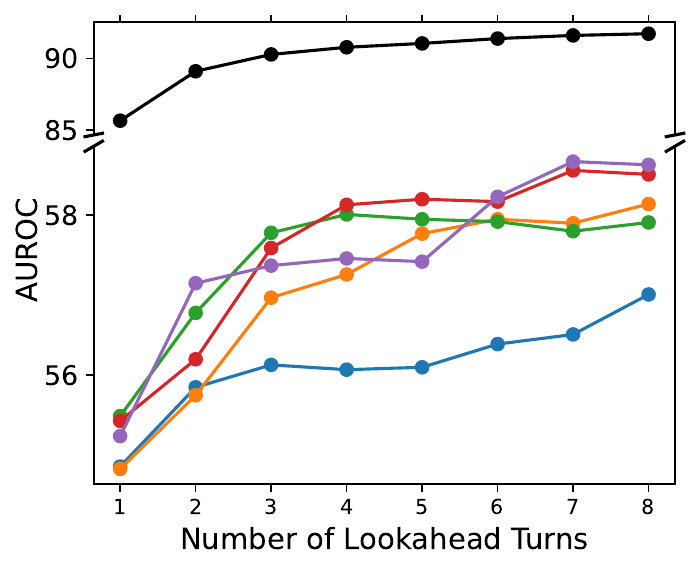}
                \caption{Detector: Gemma 3}
                \label{fig:gemma3_direct_t}
            \end{subfigure} &
            \begin{subfigure}[b]{\linewidth}
                \centering
                \includegraphics[width=\linewidth]{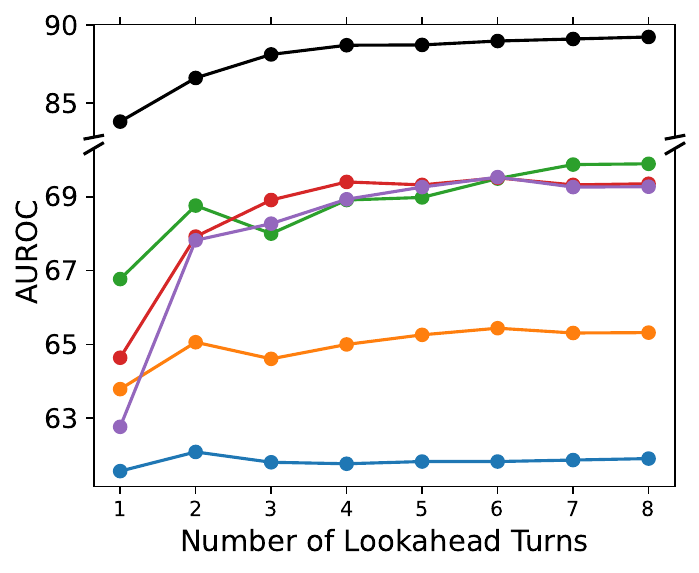}
                \caption{Detector: Qwen 2.5}
                \label{fig:qwen2.5_direct_t}
            \end{subfigure}
        \end{tabular}
    }   
    \caption{AUROC of \textsc{Direct} with future contexts.
    The first row shows performance improvements as the number of sampled future sentences from one turn ahead increases.
    The second row shows performance improvements as the number of future lookahead turns increases. 
    In both cases, incorporating more future context consistently enhances hallucination detection across all detectors.}        
    \label{fig:direct_s_t_ind}
    \vspace{-3mm}
\end{figure*}

\subsection{Analyzing the Impact of Future Sentences}
\label{sec:how_future}
We investigate which future sentence aids in detecting hallucinations in current (target) sentences.
Sampled future sentences are evaluated for hallucinations using \textsc{Direct}. 

Figure~\ref{fig:analysis_future_h} shows analysis results for cases where the future sentence is hallucinated, with an illustrative example provided in Table~\ref{tab:future_analysis_example}.
Our findings indicate that:
\textbf{(1) All Future (H) > All Future (NH)}: hallucinated current sentences are more likely to be followed by hallucinated future sentences.
\textbf{(2) Positive Future (H) > All Future (H)}: hallucinated future sentences are especially useful for detecting hallucinated current sentences.
\textbf{(3) Positive Future (NH) < All Future (NH)}: hallucinated future sentences are less useful for detecting non-hallucinated current sentences.

Figure~\ref{fig:analysis_future_nh} shows analysis results for cases where the future sentence is non-hallucinated.
Our findings indicate that:
\textbf{(1) All Future (NH) > All Future (H)}: non-hallucinated current sentences are more likely to be followed by non-hallucinated future sentences.
\textbf{(2) Positive Future (NH) > All Future (NH)}: non-hallucinated future sentences are especially useful for detecting non-hallucinated current sentences.
\textbf{(3) Positive Future (H) < All Future (H)}: non-hallucinated future sentences are less useful for detecting hallucinated current sentences.

We observed this consistent pattern across all datasets.
These findings show that future context is statistically associated with the hallucination status of the current sentence. 
Although this correlation does not establish causation, it consistently provides useful predictive signals for hallucination detection.
Consequently, future sentences are more helpful for hallucination detection when their status matches the current sentence.

\begin{table*}[t]
    \centering
    \resizebox{0.9\textwidth}{!}{%
    \begin{tabular}{ll|cccccc|c}
    \toprule
    Detector & Method & SelfCheckGPT & SC-ChatGPT & SC-GPT4 & SC-LLaMA & SC-Vicuna & True-False & Average \\
    \midrule
    \multirow{2}{*}{LLaMA 3.1}
    & \textsc{Direct}+f$_p$  & \textbf{67.5} & 68.7 & \textbf{71.2} & 66.2 & 65.1 & \textbf{91.1} & \textbf{71.6} \\
    & \textsc{Direct}+f$_n$  & \textbf{65.3} & \textbf{69.2} & 67.9 & \textbf{67.1} & 68.0 & \textbf{88.5} & 71.0 \\
    \midrule
    \multirow{2}{*}{Gemma 3}
    & \textsc{Direct}+f$_p$  & \textbf{59.1} & \textbf{58.1} & \textbf{58.6} & 58.0 & 56.3 & 88.1 & 63.0 \\
    & \textsc{Direct}+f$_n$  & 56.4 & 57.8 & 56.4 & 56.3 & 56.6 & 88.7 & 62.0 \\
    \midrule
    \multirow{2}{*}{Qwen 2.5}
    & \textsc{Direct}+f$_p$  & \textbf{63.3} & \textbf{66.4} & \textbf{70.3} & \textbf{69.3} & \textbf{69.3} & \textbf{86.4} & \textbf{70.8} \\
    & \textsc{Direct}+f$_n$  & \textbf{63.0} & 65.4 & 66.7 & \textbf{67.3} & 63.1 & 85.2 & 68.4 \\
    \bottomrule
    \end{tabular}}
    \caption{AUROC of \textsc{Direct} variants with filtered future context (+f$_p$, +f$_n$) across different LLM detectors and datasets. +f$_p$ denotes prompt-based filtering of future context, while +f$_n$ indicates NLI-based filtering. Bold indicates improvements equal to or greater than \textsc{Direct}+f.}
    \label{tab:filtered_future_context}
    \vspace{-3mm}
\end{table*}

\section{Further Findings on Future Context}
\subsection{More Future Samples Improve Performance}
\label{sec:number_of_samples}
Future context can be expanded by increasing either the lookahead turns ($t$) or samples per turn ($s$).
Figure~\ref{fig:direct_s_t_ind} (top row) shows the impact of the number of sampled future sentences on performance \((t = 1)\) in \textsc{Direct}+f.  
As the number of sampled future sentences increases, the available context expands, enabling multiple evaluations of hallucination presence.  
Moreover, sampling future sentences along multiple paths allows the influence of the target sentence to be captured in diverse ways.  
Sampling more future sentences increases helpful information, improving hallucination detection.

Additionally, Figures~\ref{fig:selfcheckgpt_s_sc_s} present results for integrating sampled future sentences into the previous methods, \textsc{SelfCheckGPT} \((w = 1)\) and \textsc{SC} \((c = 1)\).  
In all cases, increasing the number of samples leads to performance gains.  
Our method performs consistently across different detection frameworks, indicating that it is not strongly dependent on a specific approach.

\subsection{Longer Lookahead Improves Performance}
\label{sec:future_turns}
Figure~\ref{fig:direct_s_t_ind} (bottom row) presents results based on varying the lookahead turn, i.e., how far into the future the model explores.
To observe the effect of lookahead turn, the number of sampled sentences is fixed to one.
We observe that increasing lookahead turn generally leads to performance improvements, regardless of the detector.
As shown in Figure~\ref{fig:hall_stats}, when the current sentence is hallucinated, the likelihood of hallucination increases in more distant future sentences.
This suggests that the current sentence has a stronger snowballing effect on later turns, making it effective to utilize future sentences from farther ahead.

\begin{figure}[t]
    \centering
    \includegraphics[width=0.65\linewidth]{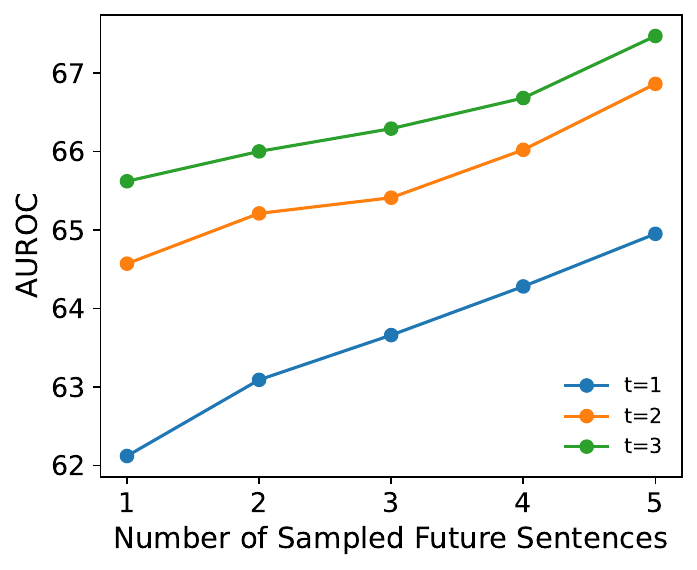}
    \caption{AUROC of \textsc{Direct} with future contexts on SelfCheckGPT dataset as both the number of sampled future sentences and the future lookahead turns increase (detector: LLaMA 3.1).}      
    \label{fig:direct_st}
    \vspace{-3mm}
\end{figure}

\subsection{Joint Impact of Sampling Count and Lookahead Turn}
\label{sec:s_and_t}
Figure~\ref{fig:direct_st} shows the effect of varying lookahead turns and samples per turn; all results are provided in Figure~\ref{fig:direct_st2}.
We hypothesize that increasing \(t\) better captures the "snowball effect"—the way the current sentence influences later ones—while increasing \(s\) uncovers multiple alternative continuations at each future turn.  
If the total number of future sentences is fixed at \(t \times s\), increasing \(t\) more clearly reveals how the hallucination status of the current sentence is propagated across future turns. On the other hand, increasing \(s\) provides diverse rephrasings for a single future turn, which makes it easier to verify the accuracy of the information. In practice, using both higher \(t\) and \(s\) offers a richer set of predictive clues and contributes to more accurate hallucination detection.

However, increasing both $t$ and $s$ can be computationally expensive. The trade-off between accuracy and efficiency should depend on the application: tasks needing thorough verification (e.g., legal or medical) may benefit from larger $t$ and $s$, while real-time or resource-limited settings (e.g., chat interfaces) may keep one parameter moderate or adopt adaptive strategies.
A detailed comparison between varying $t$ and $s$ is provided in Appendix~\ref{app:turn_sampling}.

\subsection{Filtering of Future Context}
\label{sec:filtering_future_context}
We explore two methods for selectively integrating semantically relevant future contexts into hallucination detection. The first approach explicitly prompts LLMs to use future contexts only when they are helpful, effectively inducing implicit filtering through detailed instructions~\cite{mit1,mit2}. The second approach employs an NLI model~\cite{he2021deberta} to measure semantic relevance between the current sentence and candidate future contexts, retaining those whose combined entailment and neutral scores exceed a threshold. While the optimal threshold may vary across models and datasets, we adopt a fixed value for simplicity, which in turn induces explicit filtering.

Table~\ref{tab:filtered_future_context} demonstrates that prompt-based filtering provides a simple yet effective improvement in hallucination detection performance.
Although we employed a fixed threshold for NLI-based filtering, we believe there is room for performance improvement depending on how semantic relevance is utilized.
These findings indicate that selectively incorporating meaningful future contexts is crucial for improving hallucination detection, highlighting the potential for further research on advanced context-selection methods.

\section{Conclusion}
We propose a method to detect hallucinations by sampling generated future contexts without relying on generator information.
We provides a model-agnostic benchmark and achieves improved detection performance when combined with existing methods (\textsc{SelfCheckGPT} and \textsc{SC}) and the baseline \textsc{Direct}.
Particularly, integrating future contexts with \textsc{SelfCheckGPT} effectively reduces sampling costs.
Evaluations across various turns and samples highlight the effectiveness of future contexts in sampling-based hallucination detection.

\section*{Limitations}
\label{sec:limitation}

\subsection*{Quality of Future Sentence}
\label{sec:limitation_quality}
Section~\ref{sec:filtering_future_context} discusses selective leveraging of future contexts; however, when only low-quality samples are available, its effectiveness becomes marginal. Table~\ref{tab:future_quality_s} and Table~\ref{tab:future_quality_t} analyze such cases, typically caused by insufficient generation or duplication. In particular, Qwen2.5 often produces generic duplicates (e.g., "Sure!"), and its weaker instruction-following reduces usable context, limiting the benefit of longer lookahead turns (Figure~\ref{fig:future_source}, Figure~\ref{fig:qwen2.5_direct_t}). While these issues stem from model biases and could be alleviated by prompt engineering, we kept settings consistent across detectors. 
Future contexts produced by generators, as opposed to those from detectors, tend to form more coherent paths and seldom include problematic sentences (Appendix~\ref{app:source}).

\subsection*{Correlation as a Sufficient Signal for Detection}
While our analysis in Section~\ref{sec:how_future} highlights that future sentences are statistically correlated with the hallucination status of the current sentence, we emphasize that causality is not a prerequisite for our approach. Our framework only requires future contexts to provide predictive signals, and thus they remain effective for hallucination detection even if they are correlational byproducts rather than causal effects.



\bibliography{custom}

\clearpage
\appendix
\begin{table*}[t]
    \centering
    \small
    \resizebox{0.9\textwidth}{!}{ 
    \begin{tabular}{m{2cm}|m{1.5cm}|m{10cm}}
        \toprule
        \textbf{Dataset} & \textbf{Type} & \textbf{Text} \\
        \hline
        \multirow{4}{*}{\centering SelfCheckGPT} 
        & Question & This is a Wikipedia passage about Matthew Aylmer, 1st Baron Aylmer: \\
        \cmidrule(lr){2-3}
        & Concept & Matthew Aylmer, 1st Baron Aylmer \\
        \cmidrule(lr){2-3}
        & Context & Matthew Aylmer, 1st Baron Aylmer (1708–1794) was an Irish soldier and colonial administrator. \\
        \cmidrule(lr){2-3}
        & Response & He was born in Dublin, the son of a barrister, and was educated at Trinity College, Dublin. \\
        \midrule

        \multirow{4}{*}{\centering SC-LLM} 
        & Question & Please tell me about 1990 Austrian Hockey League season \\
        \cmidrule(lr){2-3}
        & Concept & 1990 Austrian Hockey League season \\
        \cmidrule(lr){2-3}
        & Context & The 1990-91 Austrian Hockey League season was the 61st season of the Austrian Hockey League, the top level of ice hockey in Austria. \\
        \cmidrule(lr){2-3}
        & Response & EC Graz won the championship that season. \\
        \midrule

        \multirow{4}{*}{\centering True-False} 
        & Question & Please tell me about anteater \\
        \cmidrule(lr){2-3}
        & Concept & anteater \\
        \cmidrule(lr){2-3}
        & Context & - \\
        \cmidrule(lr){2-3}
        & Response & The giant anteater uses walking for locomotion. \\
        \bottomrule
    \end{tabular}
    }
    \caption{Examples of hallucination dataset}
    \label{tab:example}
\end{table*}

\begin{table*}[t]
    \centering
    \resizebox{0.8\textwidth}{!}{
    \begin{tabular}{lcccccc}
        \toprule
        Dataset & SelfCheckGPT & SC-ChatGPT & SC-GPT4 & SC-LLaMA & SC-Vicuna & True-False \\
        \midrule
        Hallucination O & 1392 & 491 & 207 & 236 & 185 & 1972 \\
        Hallucination X & 516 & 2935 & 1527 & 1532 & 878 & 2125 \\
        \bottomrule
    \end{tabular}
    }
    \caption{Statistics of dataset}
    \label{tab:dataset_stats}
\end{table*}

\section{Details of Datasets}
Table~\ref{tab:example} shows an example of the data format used in our experiments.
Since not all datasets are originally provided in the form of question, concept, context, and response, we reconstructed them to match this format.

Table~\ref{tab:dataset_stats} presents statistics of our dataset.
In SelfCheckGPT dataset, responses are generated continuously for questions sampled from the WikiBio dataset.
As a result, once a hallucination occurs, it can affect subsequent parts of the response, leading to a higher number of hallucinated cases.
In contrast, SC-LLM uses gold responses as context and generates only a single sentence as the response using the generator.
Therefore, it contains more non-hallucinated cases.
True-False dataset consists of six topics: Cities, Inventions, Chemical Elements, Animals, Companies, and Scientific Facts. 
True statements were crafted using reliable factual sources. 
False statements were mostly generated by pairing mismatched attributes from different factual entries, except for the Scientific Facts category, where ChatGPT was used to generate opposite statements.
Human annotators then reviewed and labeled each statement to ensure accuracy.
True-False dataset consists of individual sentences labeled as either true or false, without any accompanying context.

\section{LLM Prompts}
\label{app:template_prompts}

\subsection{Template for Future Context Sampling}
The following is the base prompt used for future context sampling.  
It is used to generate the sentence that directly follows the target sentence for hallucination detection.  
\label{app:template_fcs}
\begin{tcolorbox}[colback=gray!10, colframe=black!75, title=next sentence]
\small
\texttt{\{question\}}

\texttt{\{past context\}} \texttt{\{sentence\}}

next sentence:
\end{tcolorbox}
To sample $k$ future sentences, we instruct the model with the phrase "$k$ next sentences."

\subsection{Templates for Hallucination Detection}
\label{app:template_hd}
The prompt of \textsc{Direct} is as follows:
\begin{tcolorbox}[colback=gray!10, colframe=black!75, title=\textsc{Direct}]
\small
Here is a sentence about \texttt{\{concept\}}:

context: \texttt{\{past context\} \{future context\}}  

sentence: \texttt{\{sentence\}}  

Is this sentence accurate about the \texttt{\{concept\}}?

Answer Yes or No:
\end{tcolorbox}

where concept, question, past context, and sentence are provided by the dataset and are described in Section~\ref{sec:dataset}.
Future context is optional and may be selectively used depending on the experiment.
The goal of this prompt is to intuitively provide the model with clues, enabling it to determine whether the target sentence is hallucinatory based on its internal knowledge.
This prompt forms the core of our \textsc{Direct} baseline.

To implement prompt-based filtering in Table~\ref{tab:filtered_future_context}, we provide detailed instructions to use only relevant Future contexts, as shown below. To do this, we change the prompt to distinguish between past and Future contexts.

\begin{tcolorbox}[colback=gray!10, colframe=black!75, title=\textsc{Direct (Prompt-based Filtering)}]
\small
Here is a sentence about \texttt{\{concept\}}:

past context: \texttt{\{past context\}}  

future context: \texttt{\{future context\}}

The future context may or may not be relevant. Use it only if it helps you determine the accuracy of the sentence.

sentence: \texttt{\{sentence\}}  

Is this sentence accurate about the \texttt{\{concept\}}?

Answer Yes or No:
\end{tcolorbox}

We next describe the prompt design used in \textsc{SelfCheckGPT} for hallucination detection. Given a question, the LLM detector first generates multiple alternative context–response pairs using top-$k$ sampling (specifically, we set $k=50$). 
These pairs are then used as the context in the prompt, optionally supplemented with future context as follows:

\begin{tcolorbox}[colback=gray!10, colframe=black!75, title=\textsc{SelfCheckGPT}]
\small
Context: \texttt{\{context-response pairs\}} \texttt{\{future context\}}

Sentence: \texttt{\{sentence\}}

Is the sentence supported by the context above? Answer Yes or No:
\end{tcolorbox}

\textsc{SC} adopts a two-stage procedure to detect hallucinations by identifying logical inconsistencies among alternative completions. 
Before detecting hallucinations, the LLM-based detector samples alternative sentences corresponding to the current turn.
In the first stage—explanation generation—the model is prompted to describe informational conflicts between two statements: the original target sentence and a sampled alternative. 
Optionally, future context can replace the original description field (to maintain dataset consistency), providing a natural grounding for comparison.

\begin{tcolorbox}[colback=gray!10, colframe=black!75, title=\textsc{SC} (explanation)]
\small
Prompt:
Here is the start of a description about \texttt{\{concept\}}.

Then follow two statements.

Description: \texttt{\{future context\}}

Statement 1: \texttt{\{sentence\}}

Statement 2: \texttt{\{sampled sentence\}}

Please explain if the statements about {concept} are contradictory. Provide your explanation.
\end{tcolorbox}

In the second stage—contradiction detection—the model receives the explanation generated in the first stage and is prompted to make a binary decision on whether the two statements indeed contradict each other.

\begin{tcolorbox}[colback=gray!10, colframe=black!75, title=\textsc{SC} (detection)]
\small
\texttt{\{explanation\}}

Please conclude with Yes if there is a contradiction, otherwise No.

answer:
\end{tcolorbox}
This two-stage framework encourages the model to explicitly reason over the relationship between sampled variants, enhancing interpretability and providing fine-grained insight into how hallucinations are detected.

\begin{table*}[t]
    \centering
    \resizebox{0.8\textwidth}{!}{
    \begin{tabular}{l|cccccccccccc}
    \toprule
    \multirow{2}{*}{Detector} & \multicolumn{2}{c}{SelfCheckGPT} & \multicolumn{2}{c}{SC-ChatGPT} & \multicolumn{2}{c}{SC-GPT4} & \multicolumn{2}{c}{SC-LLaMA} & \multicolumn{2}{c}{SC-Vicuna} & \multicolumn{2}{c}{True-False} \\
    \cmidrule(lr){2-3} \cmidrule(lr){4-5} \cmidrule(lr){6-7} \cmidrule(lr){8-9} \cmidrule(lr){10-11} \cmidrule(lr){12-13}
     & s-avg & t-avg & s-avg & t-avg & s-avg & t-avg & s-avg & t-avg & s-avg & t-avg & s-avg & t-avg \\
    \midrule
    LLaMA 3.1 & 65.0 & 65.8 & 68.8 & 67.9 & 67.9 & 69.0 & 67.2 & 68.5 & 68.0 & 70.5 & 88.1 & 90.2 \\
    Gemma 3  & 56.3 & 56.1 & 57.9 & 57.1 & 56.1 & 57.5 & 56.7 & 57.6 & 56.3 & 57.5 & 88.8 & 90.2 \\
    Qwen 2.5 & 62.7 & 61.8 & 65.6 & 64.9 & 68.9 & 68.8 & 66.7 & 68.5 & 65.6 & 68.1 & 85.6 & 87.9 \\
    \midrule
    Average  & \textbf{61.3} & 61.2 & \textbf{64.1} & 63.3 & 64.3 & \textbf{65.1} & 63.5 & \textbf{64.9} & 63.3 & \textbf{65.4} & 87.5 & \textbf{89.4} \\
    \bottomrule
    \end{tabular}
    }
    \caption{Hallucination detection performance of \textsc{Direct} with future contexts.
    \textit{s-avg}: performance averaged across number of samples 1–8 (future turn fixed at 1); 
    \textit{t-avg}: performance averaged across future turns 1–8 (with one sample per turn).
    }
\label{tab:detector_s_t_avg}
\end{table*}

\begin{table*}[t]
    \centering
    \resizebox{\textwidth}{!}{  
    \begin{tabular}{c|ccc|ccc|ccc|ccc|ccc|ccc}  
        \toprule
        \multirow{2}{*}{Tokenizer} & \multicolumn{3}{c|}{SelfCheckGPT} & \multicolumn{3}{c|}{SC-ChatGPT} & \multicolumn{3}{c|}{SC-GPT4} & \multicolumn{3}{c|}{SC-LLaMA} & \multicolumn{3}{c|}{SC-Vicuna} & \multicolumn{3}{c}{True-False} \\  
        \cmidrule(lr){2-19}
        & $w_{tok}$ & $s_{tok}$ & $c_{tok}$ & $w_{tok}$ & $s_{tok}$ & $c_{tok}$ & $w_{tok}$ & $s_{tok}$ & $c_{tok}$ & $w_{tok}$ & $s_{tok}$ & $c_{tok}$ & $w_{tok}$ & $s_{tok}$ & $c_{tok}$ & $w_{tok}$ & $s_{tok}$ & $c_{tok}$ \\
        \midrule
        LLaMA 3.1 & 34.5 & 35.1 & 33.6 & 317.9 & 43.1 & 42.6 & 309.1 & 43.9 & 42.7 & 307.1 & 42.8 & 42.6 & 310 & 41.6 & 41.2 & 497.9 & 28.6 & 35.6 \\
        Gemma 3 & 39.2 & 23.7 & 22.9 & 510.7 & 33.5 & 34.1 & 510.6 & 34.4 & 34.3 & 510.8 & 34.2 & 34.3 & 510.4 & 34.8 & 34.8 & 510.3 & 18.6 & 20.3 \\
        Qwen 2.5 & 33.5 & 29.6 & 29.6 & 362.2 & 28.5 & 28.5 & 362.6 & 28.8 & 28.8 & 361.8 & 29.2 & 29.2 & 360.1 & 28.5 & 28.6 & 451.7 & 16.6 & 16.6 \\
        \bottomrule
    \end{tabular}
    }
    \caption{Average token counts for sampled outputs, where \( w_{tok} \) corresponds to alternative context-response pairs in \textsc{SelfCheckGPT}, \( c_{tok} \) corresponds to current sentence samples in \textsc{SC}, and \( s_{tok} \) corresponds to future contexts.
    }
    \label{tab:sampled_tokens}
\end{table*}

\begin{table*}[t]
\centering
    \resizebox{0.8\textwidth}{!}{
    \begin{tabular}{lcccccccccccc}
    \toprule
    \multirow{2}{*}{LLM} & \multicolumn{2}{c}{SelfCheckGPT} & \multicolumn{2}{c}{SC-ChatGPT} & \multicolumn{2}{c}{SC-GPT4} & \multicolumn{2}{c}{SC-LLaMA} & \multicolumn{2}{c}{SC-Vicuna} & \multicolumn{2}{c}{True-False} \\
    \cmidrule(lr){2-3}\cmidrule(lr){4-5}\cmidrule(lr){6-7}\cmidrule(lr){8-9}\cmidrule(lr){10-11}\cmidrule(lr){12-13}
     & IS & HR & IS & HR & IS & HR & IS & HR & IS & HR & IS & HR \\
    \midrule
    Llama 3.1 & 0\% & 0.4\% & 0\% & 0.1\% & 0\% & 0.2\% & 0\% & 0.3\% & 0\% & 0.2\% & 0\% & 11.6\% \\
    Gemma 3   & 0\% & 19.4\% & 0\% & 4.4\% & 0\% & 0.4\% & 0\% & 1.9\% & 0\% & 0.7\% & 100\% & 12.4\% \\
    Qwen 2.5  & 100\% & 16.8\% & 0\% & 15.2\% & 0\% & 11.8\% & 0\% & 14.2\% & 100\% & 17.7\% & 100\% & 24.7\% \\
    \bottomrule
    \end{tabular}
    }
    \caption{
    Percentage of low-quality future sentences based on overlap, corresponding to future sentences used in Figure~\ref{fig:direct_s_t_ind} (top row).
    IS (Insufficient Samples): Percentage of cases sampling fewer than half the number of requested sentences.
    HR (High Redundancy): Percentage of cases where more than 50\% of sampled sentences are duplicates.
    }
    \label{tab:future_quality_s}
\end{table*}

\begin{table*}[t]
\centering
    \resizebox{0.8\textwidth}{!}{
    \begin{tabular}{lcccccccccccc}
    \toprule
    \multirow{2}{*}{LLM} & \multicolumn{2}{c}{SelfCheckGPT} & \multicolumn{2}{c}{SC-ChatGPT} & \multicolumn{2}{c}{SC-GPT4} & \multicolumn{2}{c}{SC-LLaMA} & \multicolumn{2}{c}{SC-Vicuna} & \multicolumn{2}{c}{True-False} \\
    \cmidrule(lr){2-3}\cmidrule(lr){4-5}\cmidrule(lr){6-7}\cmidrule(lr){8-9}\cmidrule(lr){10-11}\cmidrule(lr){12-13}
     & IS & HR & IS & HR & IS & HR & IS & HR & IS & HR & IS & HR \\
    \midrule
    Llama 3.1 & 0\% & 0.1\% & 0\% & 0.1\% & 0\% & 0.1\% & 0\% & 0.1\% & 0\% & 0.6\% & 0\% & 1.2\% \\
    Gemma 3   & 0\% & 0\% & 0\% & 0\% & 0\% & 0\% & 0\% & 0\% & 0\% & 0\% & 0\% & 0\% \\
    Qwen 2.5  & 100\% & 88.2\% & 0\% & 7.4\% & 0\% & 6.1\% & 0\% & 7.2\% & 0\% & 6.5\% & 0\% & 0.6\% \\
    \bottomrule
    \end{tabular}
    }
    \caption{
    Percentage of low-quality future sentences based on overlap, corresponding to future sentences used in Figure~\ref{fig:direct_st2}.
    IS (Insufficient Samples): Percentage of cases sampling fewer than half the number of requested sentences.
    HR (High Redundancy): Percentage of cases where more than 50\% of sampled sentences are duplicates.
    }
    \label{tab:future_quality_t}
\end{table*}

\section{Lookahead Turn vs. Sampling Count}
\label{app:turn_sampling}
We find that both increasing the number of samples and extending the lookahead turn for future context sampling are effective strategies.
Increasing the number of samples allows the model to verify the influence of the current sentence across multiple possible paths,
whereas increasing the lookahead turn captures the accumulation of that influence over time.
Table~\ref{tab:detector_s_t_avg} presents a comparison between these two approaches.
We sample with both the number of samples and the number of lookahead turns ranging from 1 to 8 and report the average results.
The results indicate a trend where increasing lookahead turns provides more benefit than increasing the number of samples.
As discussed in Limitation~\ref{sec:limitation_quality}, since the quality of future sentences was not controlled, the results may vary depending on the detector and dataset.

\begin{figure}[t]
    \centering
    \includegraphics[width=0.65\linewidth]{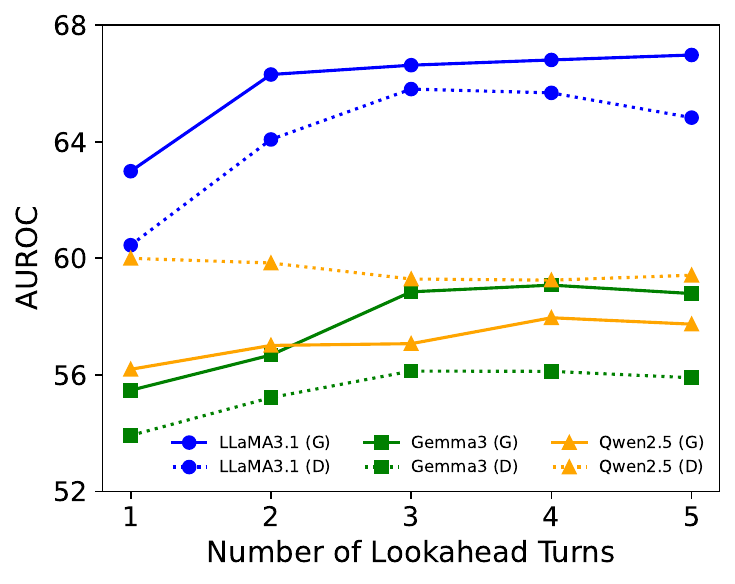}
    \caption{Hallucination detection performance comparison based on the source of sampled future context: Generator vs. Detector.}    
    \label{fig:future_source}
\end{figure}

\section{Future Context Source: Generator vs. Detector}
\label{app:source}
Figure~\ref{fig:future_source} presents experimental results based on the source of the future context.
While we generally do not have access to the generator, there are scenarios where the entire generated response is available — for instance, in blog posts or online articles that include the full text content.
In such cases, it is possible to utilize future context extracted from the generator’s own output.
SelfCheckGPT dataset provides full responses generated by the model, allowing us to use the sentences following the hallucination detection target as future context.
Since the generator provides only a single path, we also limit detector samples to one ($s = 1$).
For consistency, we evaluate only cases with at least five future sentences, allowing comparison across different $t$ values.

Future context generated by the generator (GF) leads to improved performance as the number of future turns increases.
Future context generated by the detector (DF) also contributes to performance gains for LLaMA 3.1 and Gemma 3, but shows no clear benefit for Qwen 2.5.
After analyzing the DF for Qwen 2.5, we found that the sampled future sentences in the SelfCheckGPT dataset were of relatively low quality compared to other detectors.
Further discussion on this limitation is provided in Section~\ref{sec:limitation_quality}.
Overall, while leveraging the generator’s output is preferable, sampling future context using the detector can be an effective alternative in a black-box generator setting.

\section{Aggregation of Future Sentences}
\label{app:agg_future}
When utilizing sampled future sentences, if each sentence is used as an individual future context, then $s$ sampled future sentences require $s$ separate inferences.
To reduce the number of inferences, we concatenate the sampled one-turn-ahead future sentences into a single future context and evaluate performance accordingly.
Therefore, aggregating the future sentences does not increase the number of inference steps.

\begin{figure*}[t]
    \centering
    \resizebox{1.0\textwidth}{!}{  
        \begin{tabular}{ccc}
            \begin{subfigure}{0.32\textwidth}
                \centering
                \includegraphics[width=\textwidth]{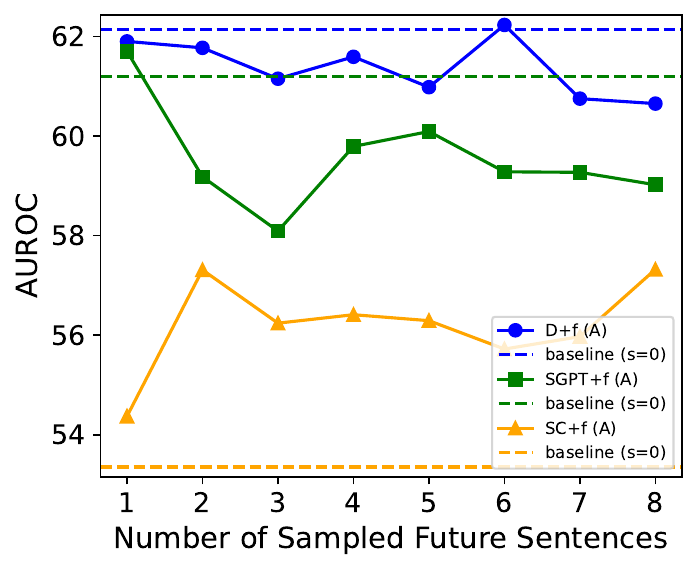}
                \caption{Dataset: SelfCheckGPT}
                \label{fig:selfcheckgpt}
            \end{subfigure} &
            \begin{subfigure}{0.32\textwidth}
                \centering
                \includegraphics[width=\textwidth]{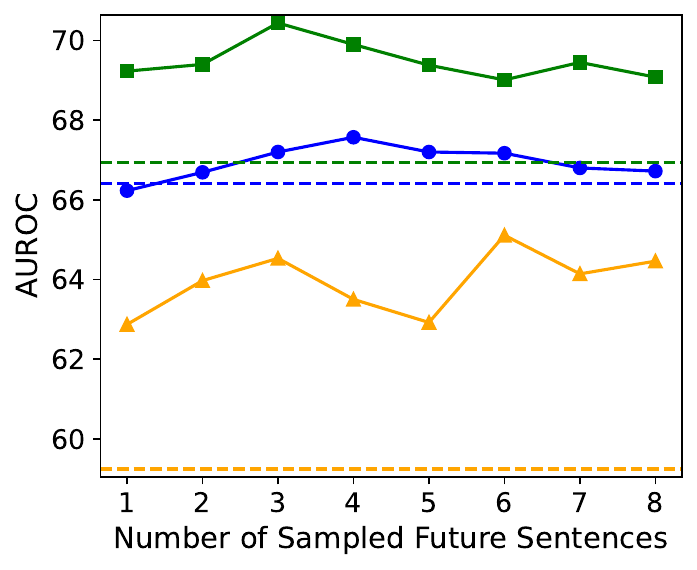}
                \caption{Dataset: SC-ChatGPT}
                \label{fig:chatgpt}
            \end{subfigure} &
            \begin{subfigure}{0.32\textwidth}
                \centering
                \includegraphics[width=\textwidth]{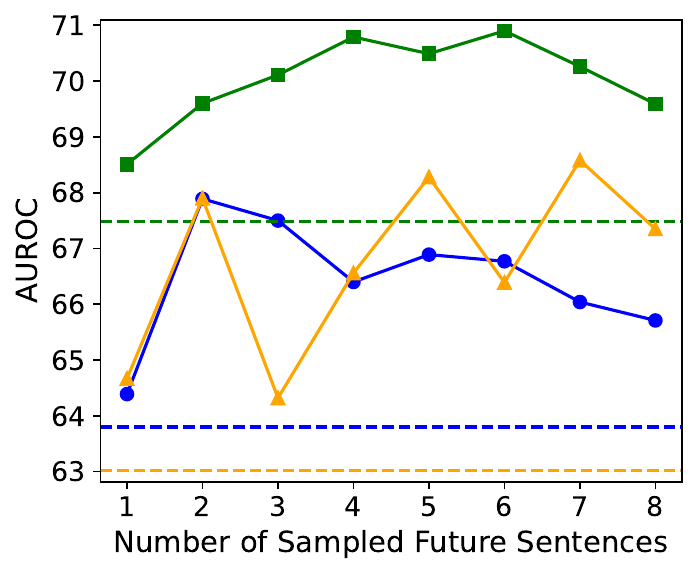}
                \caption{Dataset: SC-GPT4}
                \label{fig:gpt4}
            \end{subfigure} \\
            \begin{subfigure}{0.32\textwidth}
                \centering
                \includegraphics[width=\textwidth]{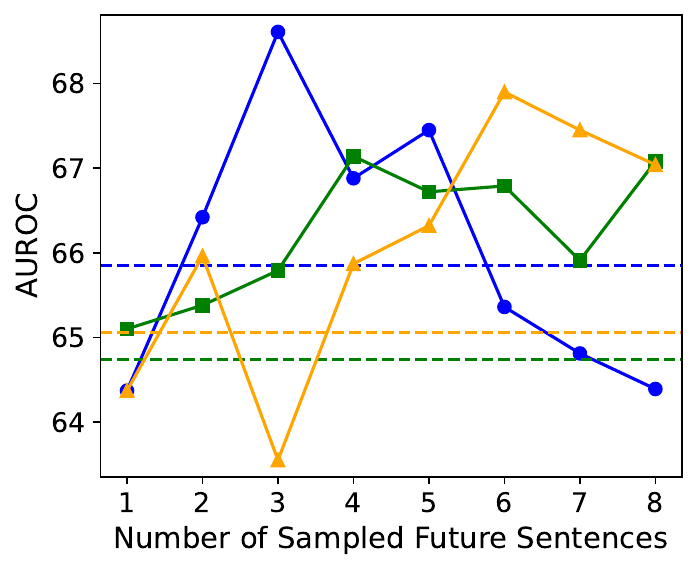}
                \caption{Dataset: SC-LLaMA}
                \label{fig:llama}
            \end{subfigure} &
            \begin{subfigure}{0.32\textwidth}
                \centering
                \includegraphics[width=\textwidth]{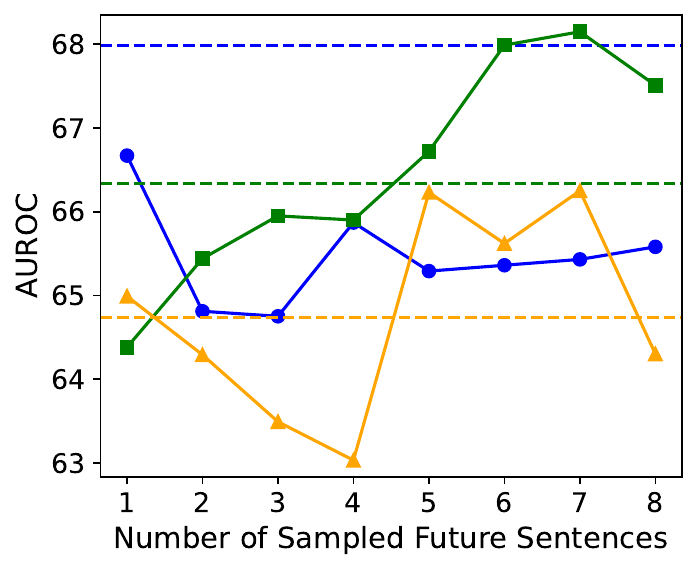}
                \caption{SC-Vicuna}
                \label{fig:vicuna}
            \end{subfigure} &
            \begin{subfigure}{0.32\textwidth}
                \centering
                \includegraphics[width=\textwidth]{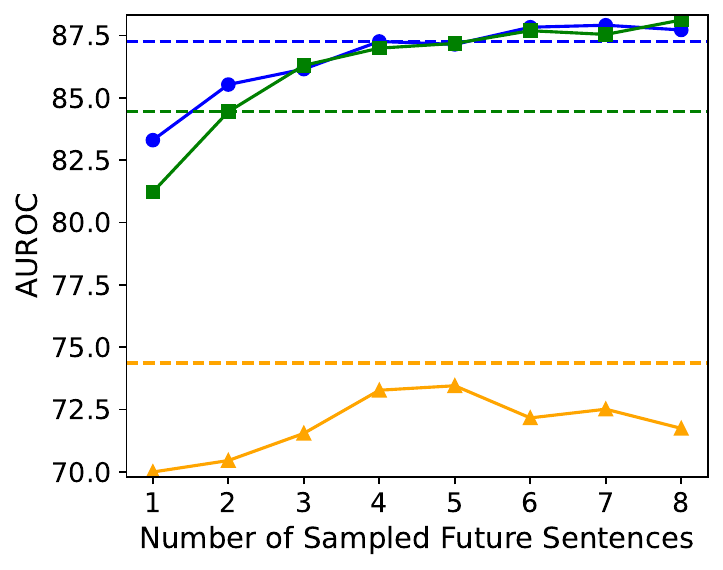}
                \caption{Dataset: True-False}
                \label{fig:true_false}
            \end{subfigure}
        \end{tabular}
    }
    \caption{Hallucination detection performance of D (\textsc{Direct}), SGPT (\textsc{SelfCheckGPT}), and \textsc{SC}. (A) represents aggregating sampled future sentences into a single future context (detector: LLaMA 3.1). The baseline is the performance without future sentences.}

    \label{fig:all_agg}
\end{figure*}

Figure~\ref{fig:all_agg} illustrate the effect of the aggregation of future sentences.
Aggregating sampled future sentences into one future context often improves performance, but occasionally may degrade it.
We believe that this is due to the presence of sampled future sentences that do not contribute to hallucination detection and may instead degrade performance.
For instance, consider a scenario where the current sentence is accurate, but the sampled future sentence contains hallucinated information or irrelevant content.
In such cases, the concatenated future context may become contaminated, providing the model with confusing clues and ultimately degrading detection performance.
A simple yet effective solution is to filter out irrelevant or hallucinated sampled future sentences, as discussed in Section \ref{sec:filtering_future_context}. This will be left as future work.

\begin{figure*}[t]
    \centering
    \begin{subfigure}{0.31\textwidth}
        \centering
        \includegraphics[width=\textwidth]{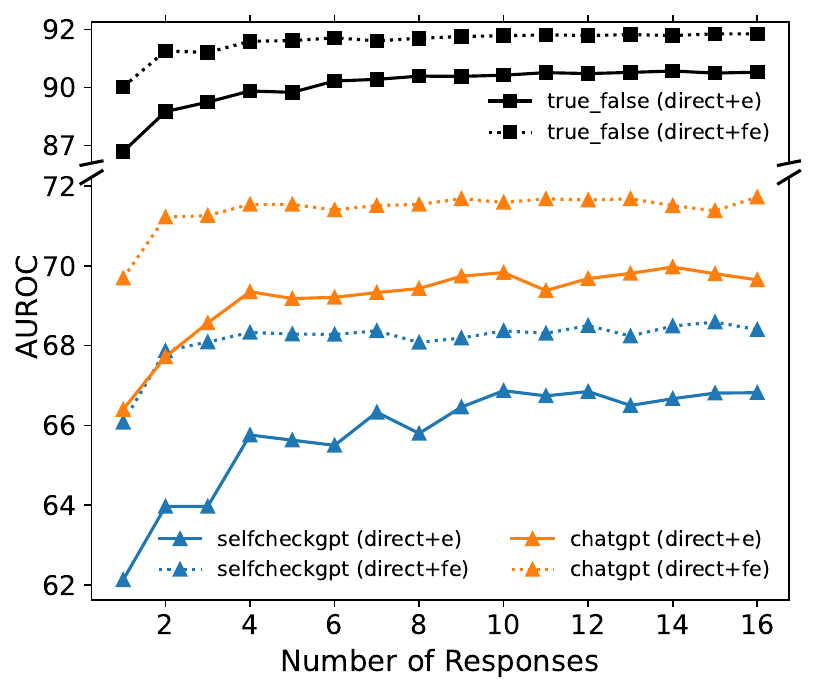}
        \caption{Detector: LLaMA 3.1 (SelfCheckGPT, SC-ChatGPT, True-False)}
        \label{fig:sampling1-llama}
    \end{subfigure}\quad
    \begin{subfigure}{0.31\textwidth}
        \centering
        \includegraphics[width=\textwidth]{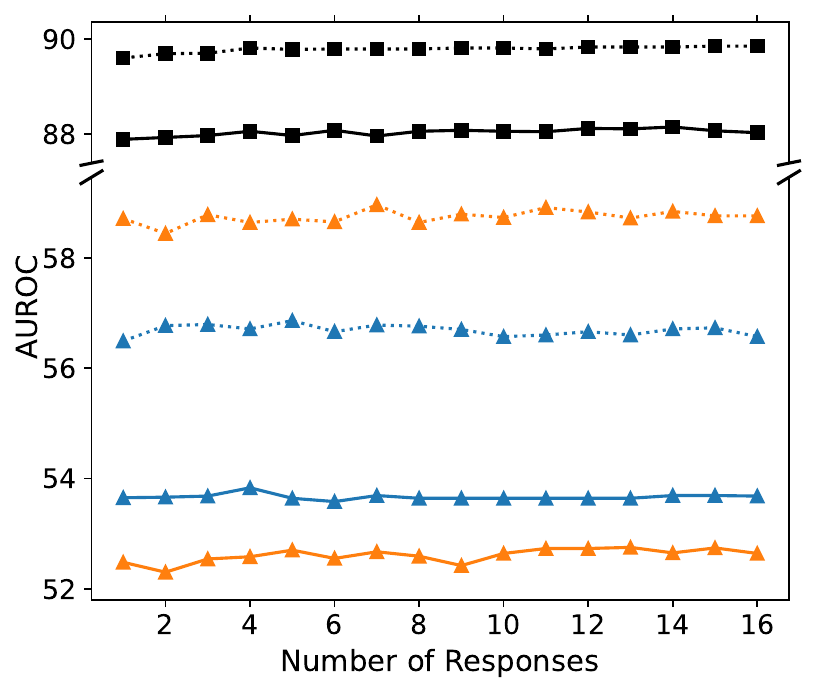}
        \caption{Detector: Gemma 3 (SelfCheckGPT, SC-ChatGPT, True-False)}
        \label{fig:sampling1-gemma}
    \end{subfigure}\quad
    \begin{subfigure}{0.31\textwidth}
        \centering
        \includegraphics[width=\textwidth]{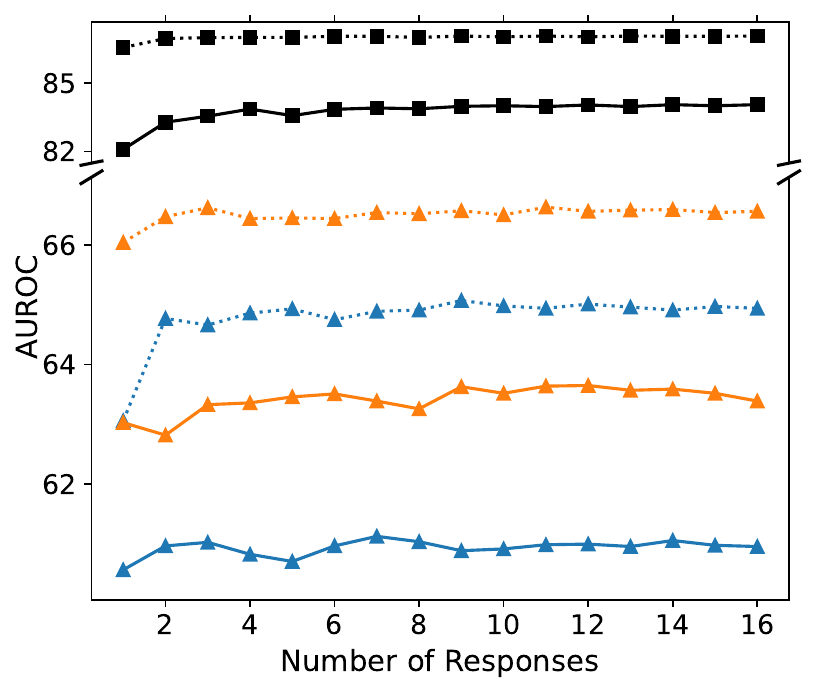}
        \caption{Detector: Qwen2.5 (SelfCheckGPT, SC-ChatGPT, True-False)}
        \label{fig:sampling1-qwen}
    \end{subfigure}

    \vspace{1em}

    \begin{subfigure}{0.31\textwidth}
        \centering
        \includegraphics[width=\textwidth]{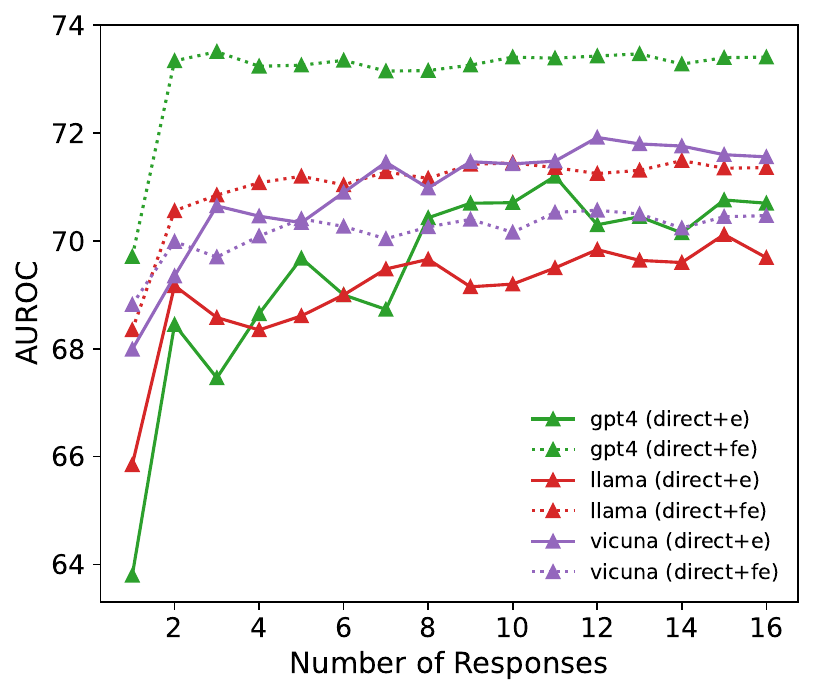}
        \caption{Detector: LLaMA 3.1 (SC-GPT4, SC-LLaMA, SC-Vicuna)}
        \label{fig:sampling2-llama}
    \end{subfigure}\quad
    \begin{subfigure}{0.31\textwidth}
        \centering
        \includegraphics[width=\textwidth]{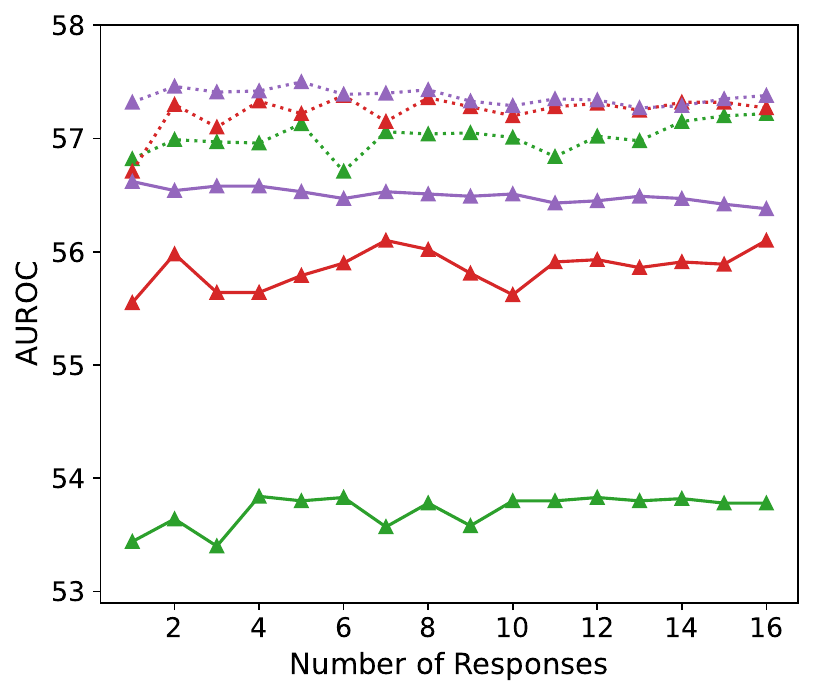}
        \caption{Detector: Gemma 3 (SC-GPT4, SC-LLaMA, SC-Vicuna)}
        \label{fig:sampling2-gemma}
    \end{subfigure}\quad    
    \begin{subfigure}{0.31\textwidth}
        \centering
        \includegraphics[width=\textwidth]{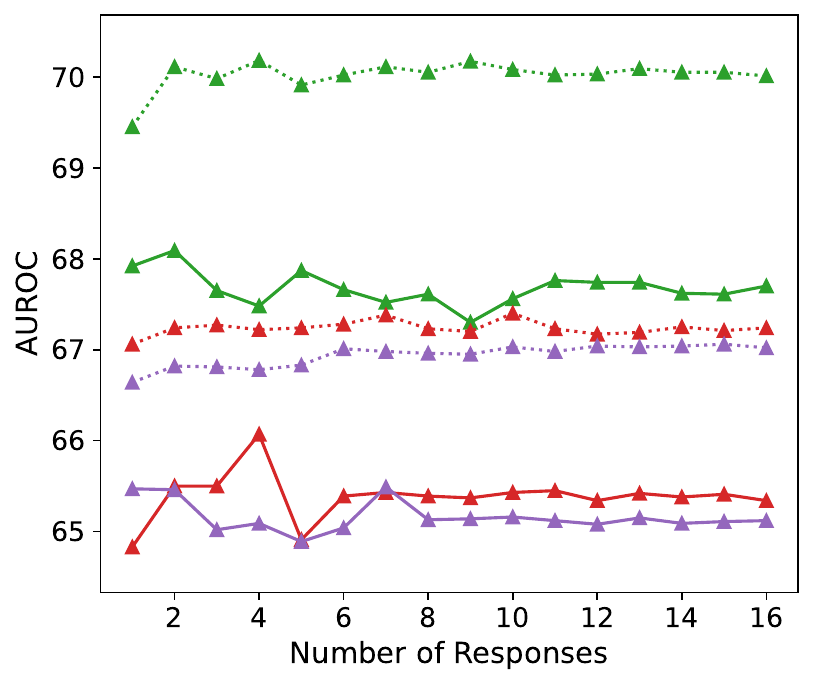}
        \caption{Detector: Qwen2.5 (SC-GPT4, SC-LLaMA, SC-Vicuna)}
        \label{fig:sampling2-qwen}
    \end{subfigure}

    \caption{Hallucination detection performance of \textsc{Direct} as the number of sampled responses increases (+e), comparing settings with and without future context (+f). Incorporating future context consistently achieves higher performance than merely increasing the number of responses, highlighting the effectiveness of leveraging future context for improved hallucination detection.}
    \label{fig:sampling_effect}
\end{figure*}

\section{Effectiveness of Future Context Beyond Self-Consistency Sampling}
\label{app:self_consistenccy}
Self-Consistency~\cite{selfconsistency} typically samples multiple reasoning paths from language models prompted via chain-of-thought (CoT) and selects the final answer based on majority voting, significantly improving accuracy over greedy decoding. Notably, it also enhances performance when simply aggregating multiple sampled responses without explicit CoT prompting, demonstrating its broader applicability.

Thus, we compared how effective future context is beyond simply increasing response sampling.
Figure~\ref{fig:sampling_effect} shows the performance obtained by sampling multiple responses (+e) directly.
\textsc{Direct}+fe uses eight sampled responses as future context.
For LLaMA 3.1, hallucination detection performance improves with more sampled responses, whereas Gemma 3 and Qwen 2.5 show minimal improvements.
However, incorporating future context (+f) consistently achieves a higher upper bound across all cases.
Even when controlling for the total number of sampled responses, incorporating future context significantly enhances performance.
For example, on the SelfCheckGPT dataset, when fixing the total number of sampled responses to 16, \textsc{Direct}+fe(2) achieves higher performance than \textsc{Direct}+e(16).

\begin{table}[t]
    \centering
    \resizebox{0.34\textwidth}{!}{%
    \begin{tabular}{lcc}
    \toprule
    Dataset & \textsc{VeriScore} & \textsc{VeriScore}+f \\
    \midrule
    True-False & 88.6 & \textbf{91.7} \\
    \bottomrule
    \end{tabular}
    }
    \caption{Hallucination detection performance (AUROC) of \textsc{VeriScore} on True-False.}
    \label{tab:rag}
\end{table}

\section{Enhancing Retrieval-based Methods with Future Context}
\label{app:retreival_methods}
Retrieval-based methods leverage external knowledge to verify factual claims and mitigate hallucinations.
~\citet{factscore} propose \textsc{\textbf{FActScore}}, a fine-grained metric for assessing factual precision by decomposing long-form text into atomic facts and verifying each fact against a reliable knowledge source.  
\citet{longfact} propose \textsc{\textbf{SAFE}}, which verifies atomic facts using retrieval-augmented LLMs with multi-step Google Search.
\citet{veriscore} introduce \textsc{\textbf{VeriScore}}, which selectively extracts context-aware, verifiable claims and verifies them through Google Search with few-shot prompting or fine-tuned open-weight models.

We also conduct a brief investigation into whether future context is beneficial in retrieval-based hallucination detection. Retrieval-based approaches leverage search results to access external knowledge, thereby benefiting from strong world knowledge. To evaluate this, we use the True-False dataset, as its sentences are structured as individual, self-contained statements that can be treated as atomic facts, making it well-suited for assessing the factuality of discrete claims in a retrieval-based setting. Table~\ref{tab:rag} presents the performance comparison between \textsc{VeriScore} and \textsc{VeriScore}+f ($s = 8$). 
\textsc{VeriScore}+f appends future context to retrieved passages from Google Search, improving performance and indicating its effectiveness for hallucination detection.
Following the main experimental setup of \citet{veriscore}, we use GPT-4o as the detector.

Retrieved passages can be incomplete, loosely related, or lack sufficient context for accurate detection. In such cases, future context sampled by the LLM provides complementary evidence, filling informational gaps missed by retrieved passages.
A more comprehensive investigation into the broader effectiveness of future context in retrieval-based hallucination detection is left for future work.

\begin{table}[t]
\centering
    \resizebox{0.3\textwidth}{!}{
    \begin{tabular}{lcc}
    \toprule
    Detector & \textsc{Direct} & \textsc{Direct}+f \\
    \midrule
    LLaMA 3.1 & 58.1 & \textbf{59.5} \\
    Gemma 3   & 51.6 & \textbf{56.5} \\
    Qwen 2.5  & 51.6 & \textbf{57.1} \\
    \bottomrule
    \end{tabular}
    }
    \caption{
    AUROC for passage-level hallucination detection.
    }
    \label{tab:passage_performance}
\end{table}

\section{Passage-Level Hallucination Detection}
\label{app:passage_level}
Following the approach used in \textsc{SelfCheckGPT}, the passage-level labels were generated by averaging sentence-level labels. The prompts for \textsc{Direct} and \textsc{Direct}+f were identical to those described in Appendix~\ref{app:template_hd}, except for replacing the term "sentence" with "passage." 
Table~\ref{tab:passage_performance} clearly demonstrates that incorporating future context also leads to performance improvements at the passage level.
Other datasets were not evaluated due to their primary focus on sentence-level hallucinations.

\begin{table*}[t]
    \centering
    \resizebox{0.9\textwidth}{!}{
    \begin{tabular}{ll|cccccc|c}
    \toprule
    \multirow{2}{*}{Detector} & \multirow{2}{*}{Method} & \multicolumn{6}{c|}{Dataset} & \multirow{2}{*}{Average} \\
    & & SelfCheckGPT & SC-ChatGPT & SC-GPT4 & SC-LLaMA & SC-Vicuna & True-False & \\    
    \midrule
    \multirow{6}{*}{LLaMA 3.1}
    & \textsc{Direct}         & 33.3 & 90.0 & 91.1 & 90.6 & 88.4 & 84.5 & 79.6 \\
    & \textsc{Direct}+f       & \textbf{36.4} & \textbf{91.5} & \textbf{92.4} & \textbf{91.5} & \textbf{89.2} & \textbf{87.1} & \textbf{81.3} \\
    \cmidrule{2-9}
    & \textsc{SelfCheckGPT}      & 42.7 & 92.0 & 94.7 & 92.2 & 90.2 & 88.1 & 83.3 \\
    & \textsc{SelfCheckGPT}+f    & \textbf{43.0} & \textbf{94.1} & \textbf{95.3} & \textbf{93.4} & \textbf{91.1} & \textbf{89.3} & \textbf{84.3} \\
    \cmidrule{2-9}
    & \textsc{SC}        & 28.6 & 89.1 & 91.7 & 91.0 & 88.1 & 71.4 & 76.6 \\
    & \textsc{SC}+f      & \textbf{31.9} & \textbf{91.4} & \textbf{94.2} & \textbf{92.5} & \textbf{90.3} & \textbf{75.8} & \textbf{79.3} \\
    \midrule
    \multirow{6}{*}{Gemma 3}
    & \textsc{Direct}         & 28.6 & 86.3 & 88.8 & 88.0 & 84.6 & 83.8 & 76.7 \\
    & \textsc{Direct}+f       & \textbf{30.1} & \textbf{87.8} & \textbf{89.5} & \textbf{88.3} & \textbf{84.9} & \textbf{84.7} & \textbf{77.6} \\
    \cmidrule{2-9}
    & \textsc{SelfCheckGPT}      & 27.4 & 91.7 & 93.8 & 92.0 & 89.3 & 88.8 & 80.5 \\
    & \textsc{SelfCheckGPT}+f    & \textbf{32.5} & \textbf{92.5} & \textbf{94.4} & \textbf{93.2} & \textbf{90.7} & \textbf{89.7} & \textbf{82.1} \\
    \cmidrule{2-9}
    & \textsc{SC}        & 28.3 & 88.6 & 90.8 & 90.1 & 87.5 & 81.2 & 77.8 \\
    & \textsc{SC}+f      & \textbf{29.6} & \textbf{90.1} & \textbf{91.7} & \textbf{90.8} & \textbf{88.5} & 80.4 & \textbf{78.5} \\
    \midrule
    \multirow{6}{*}{Qwen 2.5}
    & \textsc{Direct}         & 32.9 & 89.6 & 92.4 & 90.6 & 87.9 & 80.5 & 79.0 \\
    & \textsc{Direct}+f       & \textbf{34.6} & \textbf{90.8} & \textbf{93.0} & \textbf{91.4} & 87.7 & \textbf{83.9} & \textbf{80.2} \\
    \cmidrule{2-9}
    & \textsc{SelfCheckGPT}      & 29.6 & 90.1 & 92.6 & 90.5 & 88.5 & 86.5 & 79.6 \\
    & \textsc{SelfCheckGPT}+f    & 29.3 & \textbf{90.6} & \textbf{93.3} & \textbf{91.1} & 88.4 & 85.4 & \textbf{79.7} \\
    \cmidrule{2-9}
    & \textsc{SC}        & 27.2 & 85.9 & 88.2 & 87.2 & 84.4 & 63.9 & 72.8 \\
    & \textsc{SC}+f      & \textbf{28.9} & \textbf{88.4} & \textbf{90.6} & \textbf{89.4} & \textbf{86.0} & \textbf{64.7} & \textbf{74.7} \\
    \bottomrule
    \end{tabular}
    }
    \caption{Hallucination detection performance (AUCPR) of \textsc{Direct}, \textsc{SelfCheckGPT}, and \textsc{SC} with and without future context (+f) across different LLM detectors.
    Bold numbers indicate performance improvements when future context (+f) are used.}
    \label{tab:aucpr}
\end{table*}

\begin{figure*}[t]
    \centering
    \renewcommand{\arraystretch}{1.2}
    \setlength{\extrarowheight}{2pt}
    \resizebox{1.0\textwidth}{!}{
        \begin{tabular}{c ccc}
            \raisebox{3.5em}{\rotatebox{90}{\textsc{SelfCheckGPT}}} &
            \begin{subfigure}{0.32\textwidth}
                \centering
                \includegraphics[width=\textwidth]{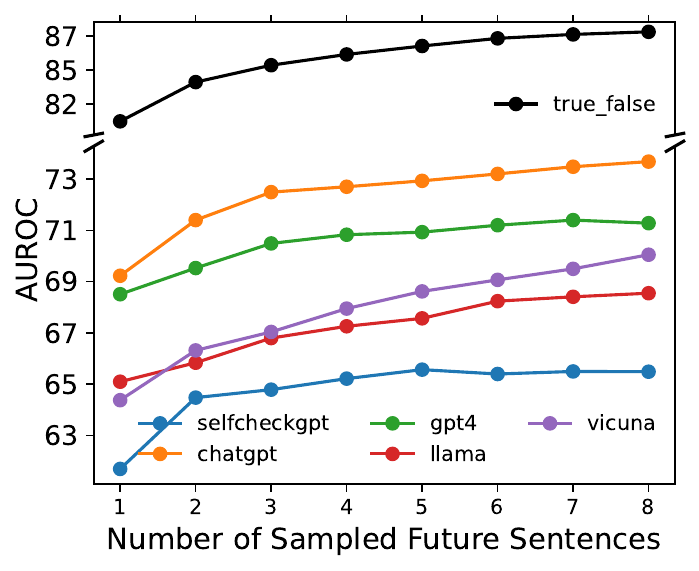}
                \caption{Detector: LLaMA 3.1}
                \label{fig:llama3.1_selfcheckgpt}
            \end{subfigure} &
            \begin{subfigure}{0.32\textwidth}
                \centering
                \includegraphics[width=\textwidth]{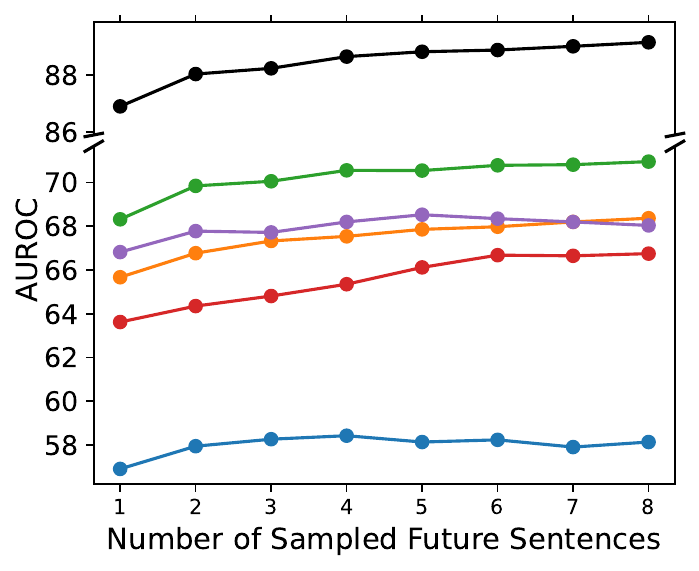}
                \caption{Detector: Gemma 3}
                \label{fig:gemma3_selfcheckgpt}
            \end{subfigure} &
            \begin{subfigure}{0.32\textwidth}
                \centering
                \includegraphics[width=\textwidth]{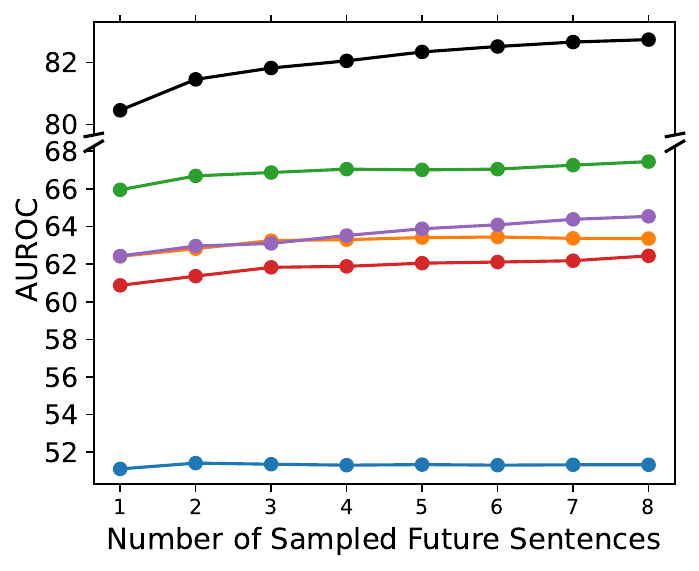}
                \caption{Detector: Qwen 2.5}
                \label{fig:qwen2.5_selfcheckgpt}
            \end{subfigure}
            \\[3ex] 
            \raisebox{7em}{\rotatebox{90}{\textsc{SC}}} & 
            \begin{subfigure}{0.32\textwidth}
                \centering
                \includegraphics[width=\textwidth]{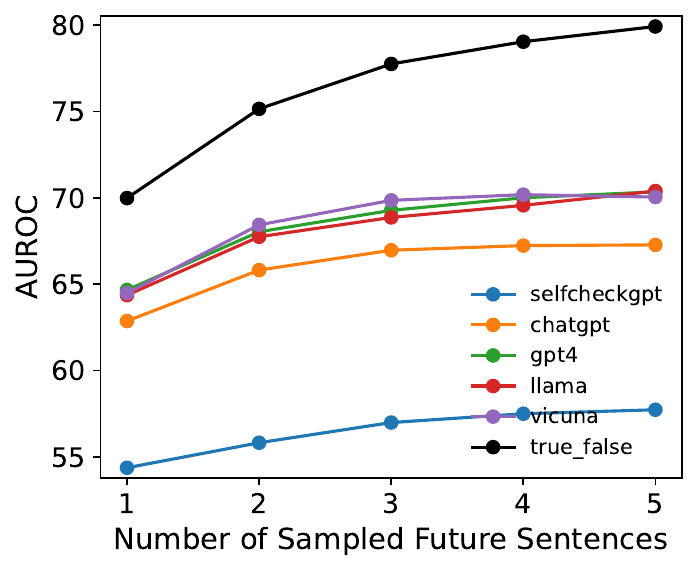}
                \caption{Detector: LLaMA 3.1}
                \label{fig:llama3.1_sc}
            \end{subfigure} &
            \begin{subfigure}{0.32\textwidth}
                \centering
                \includegraphics[width=\textwidth]{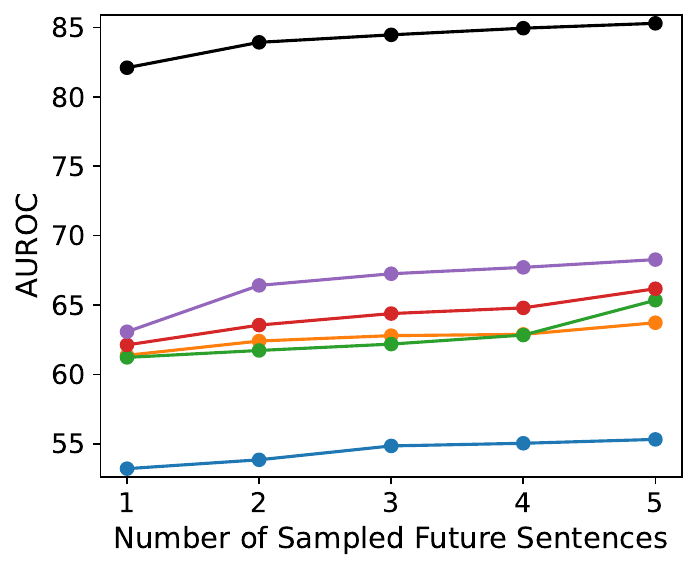}
                \caption{Detector: Gemma 3}
                \label{fig:gemma3_sc}
            \end{subfigure} &
            \begin{subfigure}{0.32\textwidth}
                \centering
                \includegraphics[width=\textwidth]{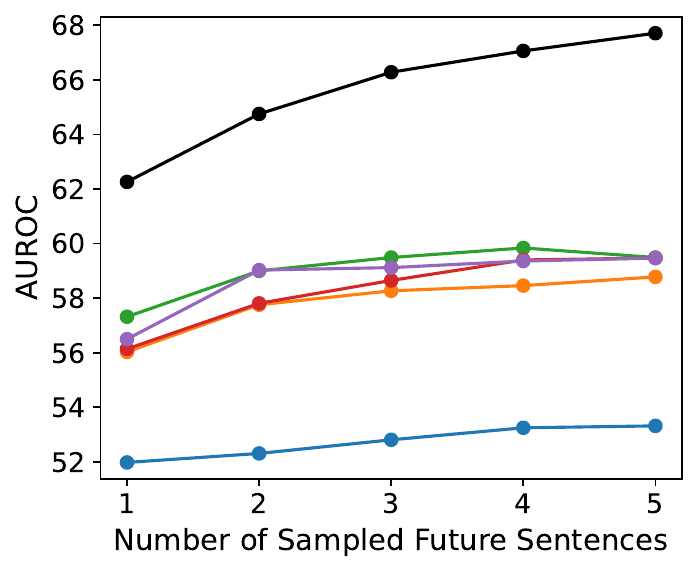}
                \caption{Detector: Qwen 2.5}
                \label{fig:qwen2.5_sc}
            \end{subfigure}
        \end{tabular}
    } 
    \caption{Hallucination detection performance of \textsc{SelfCheckGPT} and \textsc{SC} with future contexts by increasing the number of sampled future sentences from one turn ahead.
    The first row shows results for \textsc{SelfCheckGPT}, and the second row shows results for \textsc{SC}.
    Sampling more future sentences consistently improves performance across all detectors.}      
    \label{fig:selfcheckgpt_s_sc_s}
\end{figure*}

\begin{figure*}[t]
    \centering
    \renewcommand{\arraystretch}{1.2}
    \setlength{\extrarowheight}{2pt}
    \resizebox{1.0\textwidth}{!}{ 
        \begin{tabular}{c ccc}
            \raisebox{3.5em}{\rotatebox{90}{\textsc{SelfCheckGPT}}} &
            \begin{subfigure}{0.32\textwidth}
                \centering
                \includegraphics[width=\textwidth]{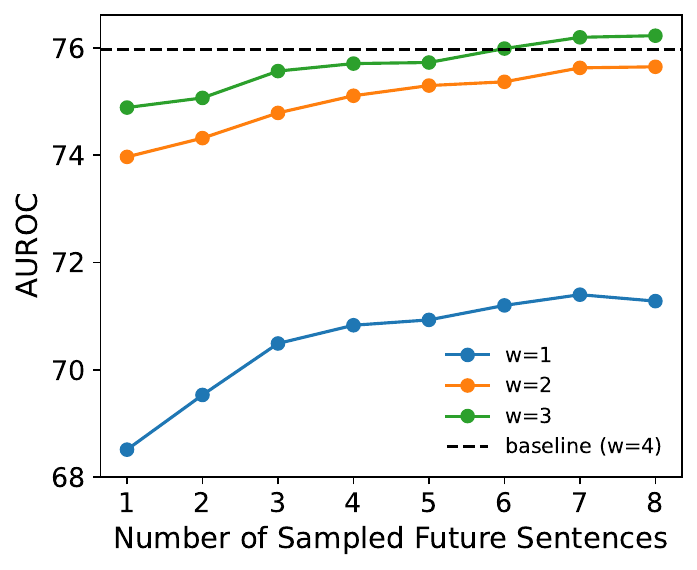}
                \caption{Dataset: SC-GPT4}
                \label{fig:selfcheckgpt_gpt4}
            \end{subfigure} &
            \begin{subfigure}{0.32\textwidth}
                \centering
                \includegraphics[width=\textwidth]{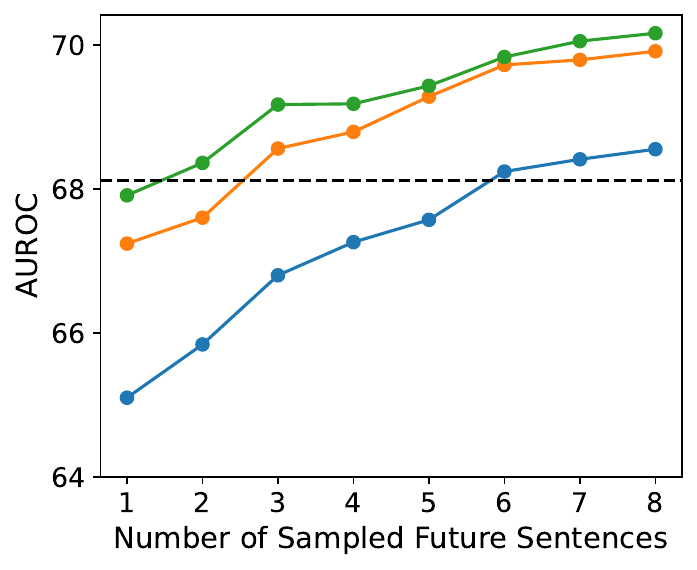}
                \caption{Dataset: SC-LLaMA}
                \label{fig:selfcheckgpt_llama}
            \end{subfigure} &
            \begin{subfigure}{0.32\textwidth}
                \centering
                \includegraphics[width=\textwidth]{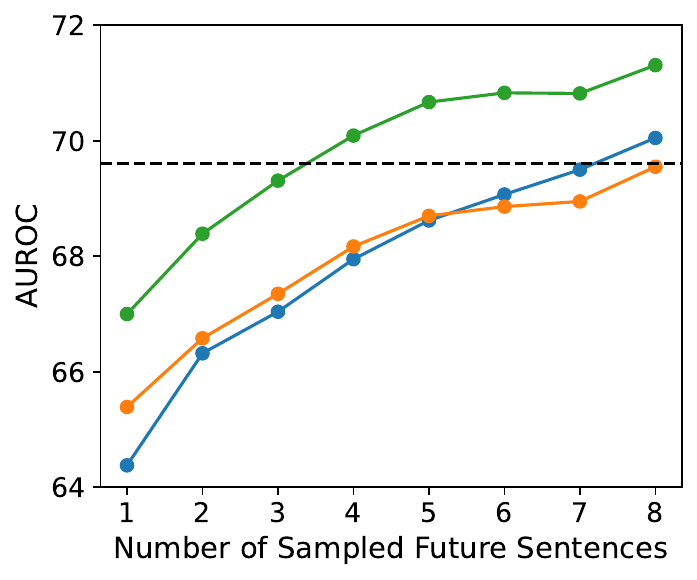}
                \caption{Dataset: SC-Vicuna}
                \label{fig:selfcheckgpt_vicuna}
            \end{subfigure}
            \\[3ex] 
            \raisebox{7em}{\rotatebox{90}{\textsc{SC}}} &
            \begin{subfigure}{0.32\textwidth}
                \centering
                \includegraphics[width=\textwidth]{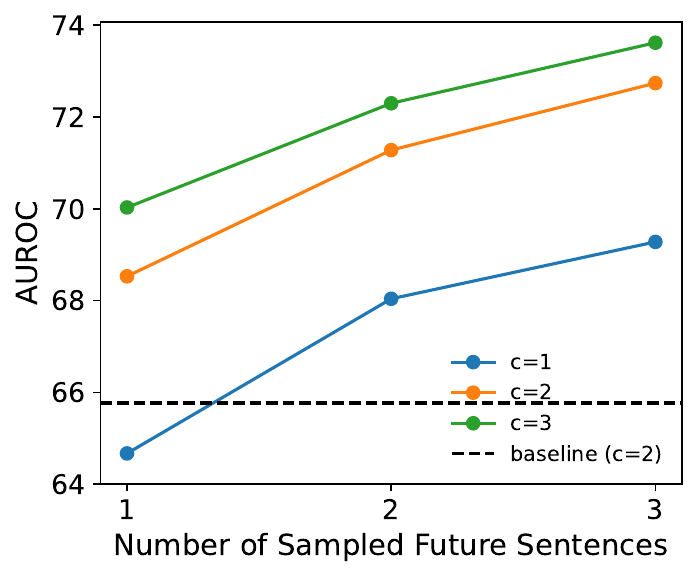}
                \caption{Dataset: SC-GPT4}
                \label{fig:sc_gpt4}
            \end{subfigure} &
            \begin{subfigure}{0.32\textwidth}
                \centering
                \includegraphics[width=\textwidth]{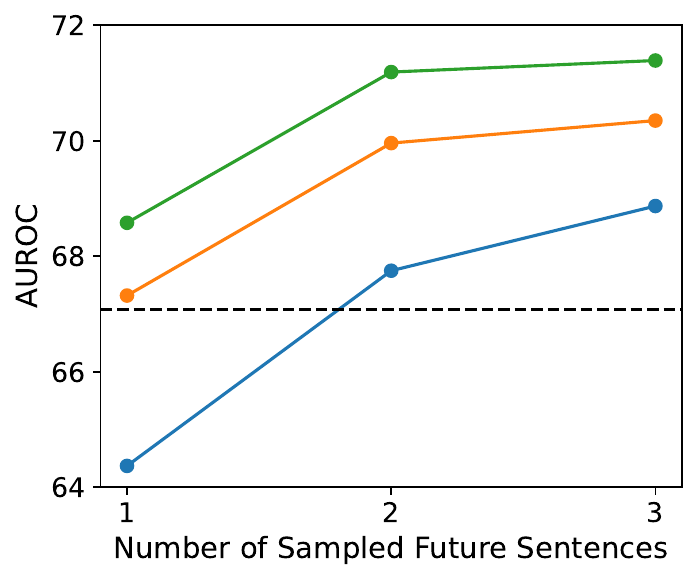}
                \caption{Dataset: SC-LLaMA}
                \label{fig:sc_llama}
            \end{subfigure} &
            \begin{subfigure}{0.32\textwidth}
                \centering
                \includegraphics[width=\textwidth]{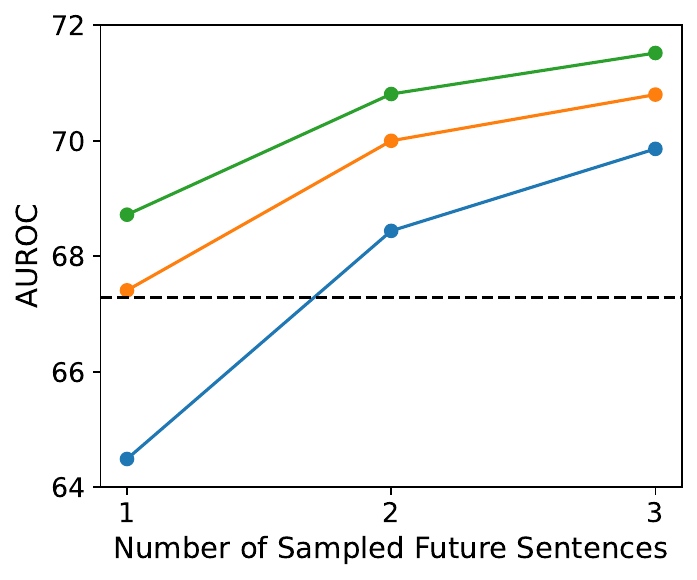}
                \caption{Dataset: SC-Vicuna}
                \label{fig:sc_vicuna}
            \end{subfigure}
        \end{tabular}
    }
    \caption{Hallucination detection performance of \textsc{SelfCheckGPT} and \textsc{SC} with future contexts (detector: LLaMA 3.1).
    The first and second rows show performance improvements for \textsc{SelfCheckGPT} and \textsc{SC}, respectively, when incorporating future contexts.
    In both cases, performance further increases as the number of sampled future contexts grows.
    }
    \label{fig:selfcheck_w_sc_c2}
\end{figure*}

\begin{figure*}[t]
    \centering
    \resizebox{1.0\textwidth}{!}{  
        \begin{tabular}{ccc}
            \begin{subfigure}{0.32\textwidth}
                \centering
                \includegraphics[width=\textwidth]{picture/direct_st/ts_selfcheckgpt.pdf}
                \caption{Dataset: SelfCheckGPT}
                \label{fig:selfcheckgpt}
            \end{subfigure} &
            \begin{subfigure}{0.32\textwidth}
                \centering
                \includegraphics[width=\textwidth]{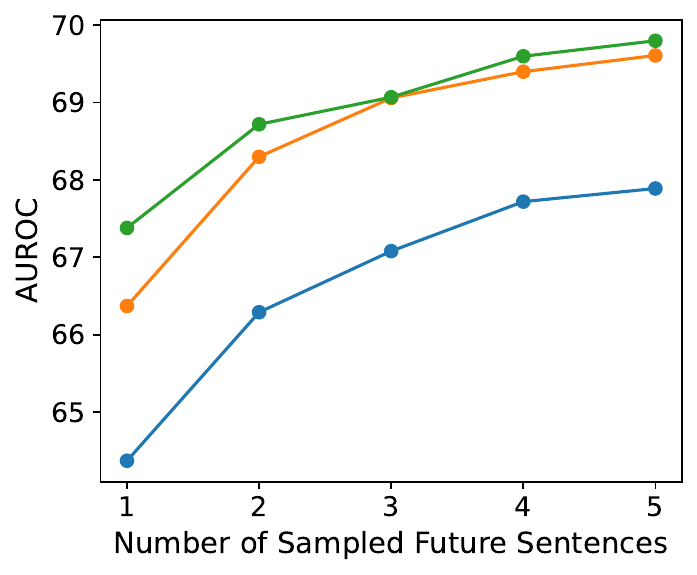}
                \caption{Dataset: SC-ChatGPT}
                \label{fig:chatgpt}
            \end{subfigure} &            
            \begin{subfigure}{0.32\textwidth}
                \centering
                \includegraphics[width=\textwidth]{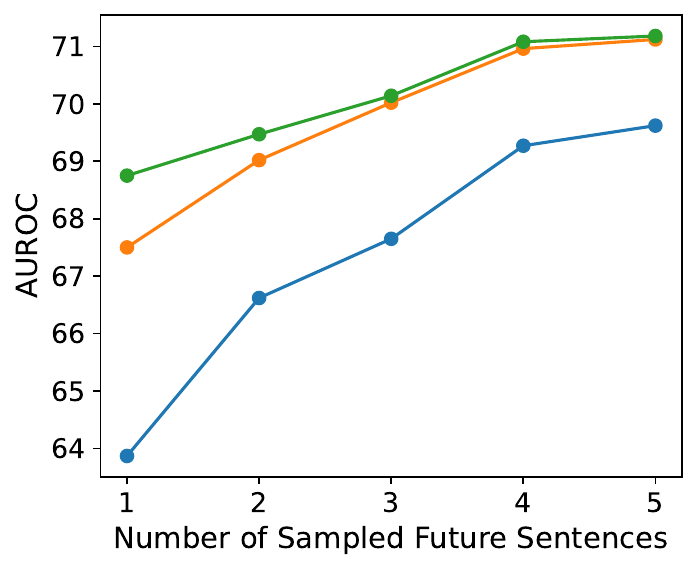}
                \caption{Dataset: SC-GPT4}
                \label{fig:gpt4}
            \end{subfigure} \\
            \begin{subfigure}{0.32\textwidth}
                \centering
                \includegraphics[width=\textwidth]{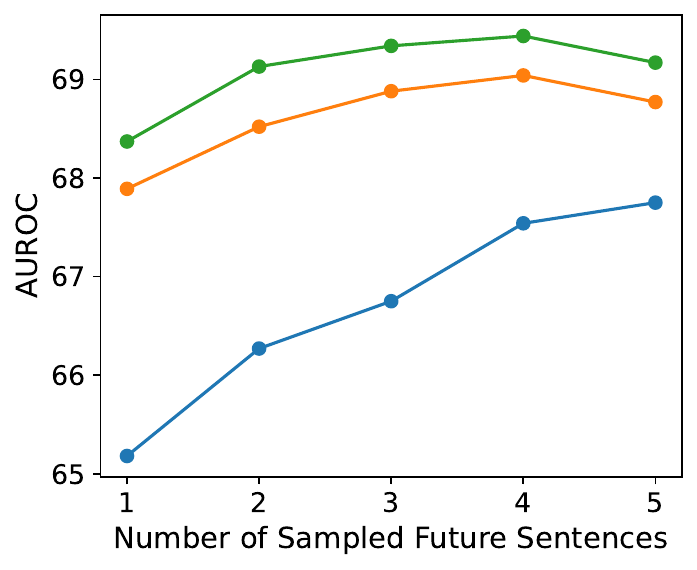}
                \caption{Dataset: SC-LLaMA}
                \label{fig:llama}
            \end{subfigure} &
            \begin{subfigure}{0.32\textwidth}
                \centering
                \includegraphics[width=\textwidth]{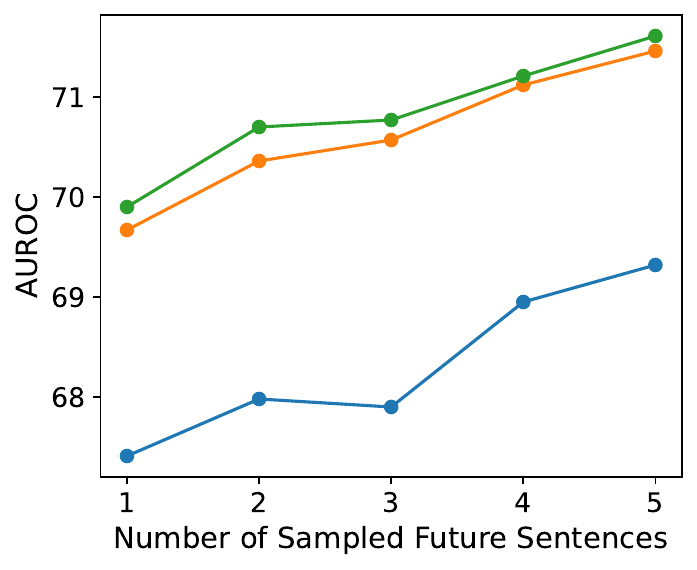}
                \caption{Dataset: SC-Vicuna}
                \label{fig:vicuna}
            \end{subfigure} &
            \begin{subfigure}{0.32\textwidth}
                \centering
                \includegraphics[width=\textwidth]{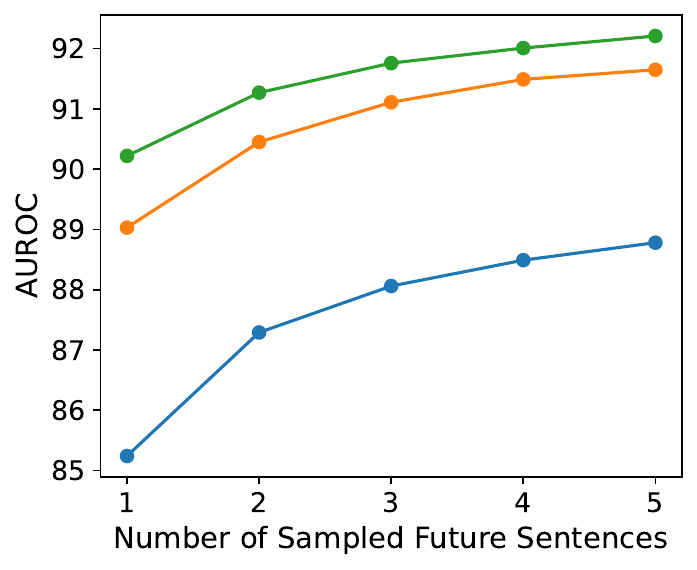}
                \caption{Dataset: True-False}
                \label{fig:true_false}
            \end{subfigure}
        \end{tabular}
    }
    \caption{Hallucination detection performance of \textsc{Direct} with future contexts as both the number of sampled future sentences and the future lookahead turns increase (detector: LLaMA 3.1).
    Performance consistently improves as both the number of sampled sentences and the lookahead turns increase.}
    \label{fig:direct_st2}
\end{figure*}

\begin{table*}[t]    
    \centering
    \resizebox{0.75\textwidth}{!}{ 
    \begin{tabular}{llll}
    \toprule
    \textbf{Current Sentence} & \textbf{Future Sentence} & \textbf{Predicted Label} & \textbf{Correct with Future Sentence?} \\
    \midrule
    \multirow{11}{*}{\makecell{Hallucinated Sentence \\ (H)}} 
    & Sentence 1 & H  & Yes \\
    & Sentence 2 & NH & Yes  \\
    & Sentence 3 & H  & Yes \\
    & Sentence 4 & NH & No  \\
    & Sentence 5 & H  & Yes \\
    & Sentence 6 & NH & No  \\
    & Sentence 7 & H  & No  \\
    & Sentence 8 & H  & Yes \\
    & Sentence 9 & H  & Yes \\
    & Sentence 10 & NH & No  \\
    \cmidrule(lr){2-4}
    & \multicolumn{3}{l}{All Future (H) = 6/10,\quad Positive Future (H) = 5/6} \\
    \midrule
    \multirow{11}{*}{\makecell{Non-Hallucinated Sentence \\ (NH)}} 
    & Sentence 1 & NH & Yes \\
    & Sentence 2 & NH & No  \\
    & Sentence 3 & NH & No  \\
    & Sentence 4 & H  & No  \\
    & Sentence 5 & NH & Yes \\
    & Sentence 6 & NH & No  \\
    & Sentence 7 & NH & No  \\
    & Sentence 8 & H  & Yes \\
    & Sentence 9 & NH & No  \\
    & Sentence 10 & NH & No  \\
    \cmidrule(lr){2-4}
    & \multicolumn{3}{l}{All Future (NH) = 2/10,\quad Positive Future (NH) = 1/2} \\
    \bottomrule
    \end{tabular}
    }
    \caption{
    Illustrative example of hallucination propagation corresponding to Figure~\ref{fig:analysis_future_h}, with 10 future sentences. 
    H and NH denote hallucinated and non-hallucinated labels, respectively.
    }    
    \label{tab:future_analysis_example}
\end{table*}

\end{document}